\documentclass[]{ailab}

\usepackage{wrapfig}
\usepackage{tabularx}
\usepackage{textcomp}
\usepackage{stfloats}
\usepackage{url}
\usepackage{verbatim}
\usepackage{titlesec}
\usepackage{tocloft}
\usepackage{adjustbox}
\usepackage{multirow}
\usepackage{pifont}
\usepackage{tikz}
\usepackage{comment}
\usepackage{amsmath,amssymb} 
\usepackage{colortbl}  
\usepackage{color}
\usepackage{booktabs} 
\usepackage{natbib} 
\setcitestyle{square,comma,numbers,sort&compress}
\usepackage{graphicx}    
\usepackage{subcaption} 
\usepackage{multirow} 
\usepackage{tcolorbox}
\usepackage{booktabs} 
\usepackage{subcaption} 
\usepackage[dvipsnames]{xcolor}
\usepackage{graphicx}
\usepackage{amsmath, amssymb}
\RequirePackage{xspace}
\makeatletter
\DeclareRobustCommand\onedot{\futurelet\@let@token\@onedot}
\def\@onedot{\ifx\@let@token.\else.\null\fi\xspace}

\usepackage{lineno}

\definecolor{darkblue}{rgb}{0, 0, 0.5}
\hypersetup{colorlinks=true, citecolor=darkblue, linkcolor=darkblue, urlcolor=darkblue}

\usepackage{wrapfig}  
\usepackage{lipsum}  

\newcommand{\mybold}[1]{\noindent\textbf{#1}}
\newcommand{\notebook}[0]{world knowledge }

\newtcbtheorem[number within=section]{exmp}{Prompts}%
{breakable,colback=white!5!white,colframe=black!95!,fonttitle=\bfseries, left=.02in, right=.02in,bottom=.02in, top=.02in}{exmp}

\newtcbtheorem[number within=section]{case_data}{Examples}%
{breakable,
 colback=white!5!white,
 colframe={rgb,255:red,78; green,136; blue,114}, 
 fonttitle=\bfseries,
 left=.02in,
 right=.02in,
 bottom=.02in,
 top=.02in}{case_data}

\definecolor{mygreen}{HTML}{85A490}
\definecolor{myred}{HTML}{C75841}

\definecolor{cotred}{RGB}{200, 30, 30}
\definecolor{ForestGreen}{RGB}{34, 139, 34}

\usepackage[normalem]{ulem}

\makeatother

\definecolor{adptorange}{RGB}{248, 205, 172}
\definecolor{cmpblue}{RGB}{189, 215, 238}
\definecolor{cmpblue}{RGB}{189, 215, 238}

\definecolor{our_red}{RGB}{232,157,160}
\definecolor{our_blue}{RGB}{136,206,230}
\definecolor{our_orange}{RGB}{246,200,168}
\definecolor{our_green}{RGB}{178,211,164}

\definecolor{attn_code0}{RGB}{247,215,200}
\definecolor{attn_code1}{RGB}{238,169,139}
\definecolor{mlp_code0}{RGB}{204,201,221}
\definecolor{mlp_code1}{RGB}{102,95,153}

\definecolor{token_blue}{RGB}{84, 120, 140}

\usepackage{amsmath,amsfonts,bm}

\def\eqref#1{equation~\ref{#1}}

\def\1{\bm{1}}

\DeclareMathAlphabet{\mathsfit}{\encodingdefault}{\sfdefault}{m}{sl}
\SetMathAlphabet{\mathsfit}{bold}{\encodingdefault}{\sfdefault}{bx}{n}

\usepackage{amsmath,amsfonts,bm}

\def\eqref#1{equation~\ref{#1}}

\def\1{\bm{1}}

\DeclareMathAlphabet{\mathsfit}{\encodingdefault}{\sfdefault}{m}{sl}
\SetMathAlphabet{\mathsfit}{bold}{\encodingdefault}{\sfdefault}{bx}{n}

\usepackage{multirow}
\usepackage{diagbox}
\usepackage{makecell}
\usepackage{tabularx}
\usepackage{graphicx}

\usepackage{array}
\usepackage{rotating}

\definecolor{aliceblue}{rgb}{0.94, 0.97, 1.0}
\definecolor{citecolor}{HTML}{0071BC}
\definecolor{linkcolor}{HTML}{ED1C24}
\definecolor{darkgreen}{HTML}{539165}

\makeatletter
\newcommand{\thickhline}{%
 \noalign {\ifnum 0=`}\fi \hrule height 1pt
 \futurelet \reserved@a \@xhline
}
\makeatother

\usepackage{pifont}

\usepackage{pifont}       
\usepackage{bbding}       
\usepackage{fontawesome}
\usepackage{xspace}
\usepackage{float}
\usepackage{enumitem}

\newlength\savewidth

\newcolumntype{x}[1]{>{\centering\arraybackslash}p{#1pt}}
\newcolumntype{y}[1]{>{\raggedright\arraybackslash}p{#1pt}}
\newcolumntype{z}[1]{>{\raggedleft\arraybackslash}p{#1pt}}

\renewcommand{\paragraph}[1]{\vspace{1mm}\noindent\textbf{#1}}
\usepackage{colortbl}
\usepackage{xcolor}
\usepackage{wrapfig}

\renewcommand{\paragraph}[1]{\vspace{1.25mm}\noindent\textbf{#1}}

\usepackage{algorithm}
\usepackage{listings}

\definecolor{codeblue}{rgb}{0.25, 0.5, 0.5}
\definecolor{codekw}{rgb}{0.35, 0.35, 0.75}
\lstdefinestyle{Pytorch}{
    language = Python,
    backgroundcolor = \color{white},
    basicstyle = \fontsize{9pt}{8pt}\selectfont\ttfamily\bfseries,
    columns = fullflexible,
    aboveskip=1pt,
    belowskip=1pt,
    breaklines = true,
    captionpos = b,
    commentstyle = \color{codeblue},
    keywordstyle = \color{codekw},
}

\definecolor{colSubject}{HTML}{D32F2F}   
\definecolor{colAction}{HTML}{F57C00}    
\definecolor{colDetail}{HTML}{388E3C}    
\definecolor{colSpatial}{HTML}{1976D2}   
\definecolor{colMood}{HTML}{7B1FA2}      
\definecolor{colKnow}{HTML}{AFB42B}      

\usepackage{xcolor}

\usepackage{tikz}
\usetikzlibrary{tikzmark, calc, shadows.blur, shapes.geometric, fit, positioning}
\usepackage{xparse}
\pgfdeclarelayer{background}
\pgfdeclarelayer{floatbox}
\pgfdeclarelayer{foreground}
\pgfsetlayers{background,floatbox,main,foreground}

\newcounter{hcellcount}

\NewDocumentCommand{\hctext}{m}{\csname hctext@#1\endcsname}
\NewDocumentCommand{\sethctext}{mm}{\expandafter\gdef\csname hctext@#1\endcsname{#2}}

\definecolor{scoreRed}{RGB}{200, 0, 0}
\definecolor{grayText}{RGB}{120, 120, 120}

\definecolor{green}{HTML}{009000}
\definecolor{red}{HTML}{ea4335}

\definecolor{cvblue}{rgb}{0.15, 0.45, 0.68}

\title{\centering Training LLM Agents for Spontaneous, Reward-Free Self-Evolution via World Knowledge Exploration}

\author[1, 2, *]{Qifan Zhang}
\author[1, *]{Dongyang Ma}
\author[1]{Tianqing Fang}
\author[2]{Jia Li}
\author[2]{Jing Tang}
\author[1]{Nuo Chen}
\author[1]{Haitao Mi}
\author[1, *, \dagger]{\newline Yan Wang}

\affiliation[1]{Tencent\\}
\affiliation[2]{The Hong Kong University of Science and Technology (Guangzhou)}
\contribution[*]{Equal contribution}
\contribution[\dagger]{Project Lead}

\email{bklight999@gmail.com, yanwang.branden@gmail.com }

\abstract{
Most agents today ``self-evolve'' by following rewards and rules defined by humans. However, this process remains fundamentally dependent on external supervision; without human guidance, the evolution stops. In this work, we train agents to possess an intrinsic meta-evolution capability to spontaneously learn about unseen environments prior to task execution. 
To instill this ability, we design an outcome-based reward mechanism that measures how much an agent's self-generated world knowledge improves its success rate on downstream tasks. This reward signal is used exclusively during the training phase to teach the model how to explore and summarize effectively.
At inference time, the agent requires no external rewards or human instructions. It spontaneously performs native self-evolution to adapt to unknown environments using its internal parameters. When applied to Qwen3-30B and Seed-OSS-36B, this shift to native evolution yields a 20\% performance increase on WebVoyager and WebWalker. Most strikingly, the generated world knowledge even enables a compact 14B Qwen3 model to outperform the unassisted Gemini-2.5-Flash, establishing a new paradigm for truly evolving agents.
}

\date{\today} 
\headercontent{
    \begin{tabular}{ccc}
        \href{https://github.com/Bklight999/world-knowledge}{\faGithub\ Code} &
        \hspace{2em}
        
        \href{https://huggingface.co/Bklight999/World-Knowledge}{%
            \raisebox{-0.15\height}{\includegraphics[height=1.1em]{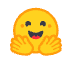}}\ Models%
        } & 
        \hspace{2em}
        
        \href{https://huggingface.co/Bklight999/World-Knowledge}{%
            \faDatabase\ Data
        }
    \end{tabular}
}

\begin{document}
\thispagestyle{firstheader}
\maketitle

\begin{figure}[t]
\centering
\includegraphics[width=\linewidth]{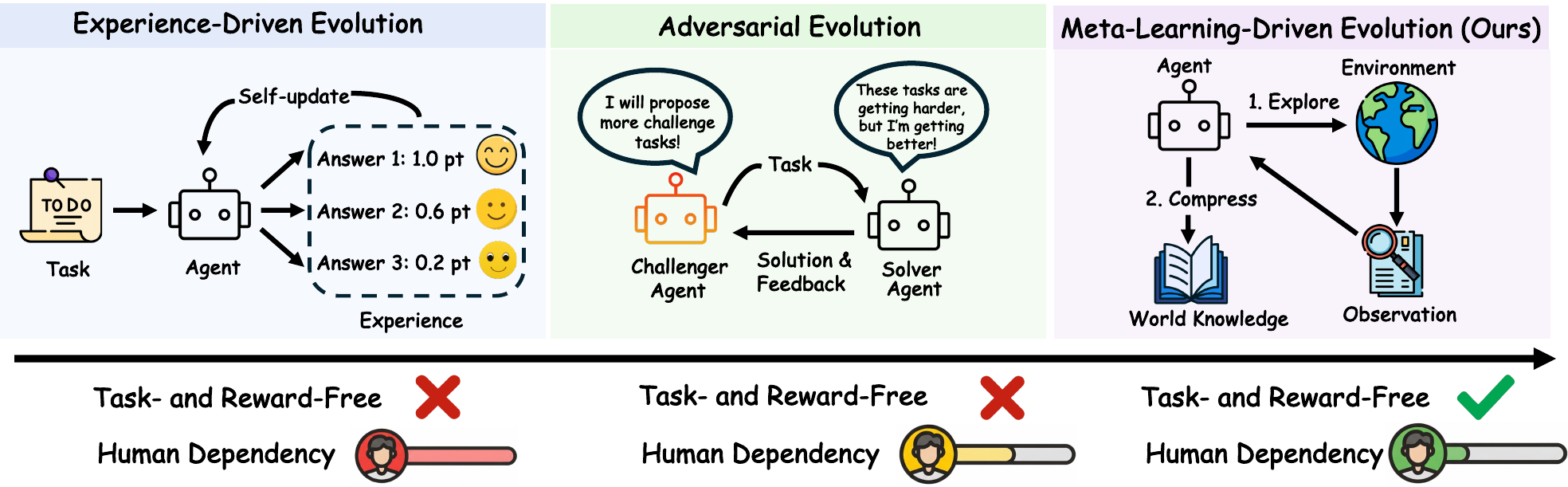}
\caption{Progression of self-evolution agent paradigms. \textbf{Left:} \textit{Experience-Driven Evolution} updates agents through predefined tasks and external rewards, requiring extensive human effort to design these components. \textbf{Center:} \textit{Adversarial Evolution} employs a challenger-solver dynamic where one proposes harder tasks and the other improves to solve them; while tasks and rewards are agent-generated, the cooperative pipeline still requires human setup. \textbf{Right:} Our \textit{Meta-Learning-Driven Evolution} enables agents to autonomously explore and compress environments into reusable world knowledge for adaptation, achieving a task- and reward-free paradigm with minimal human intervention.}
\label{fig:intro}
\end{figure}

\section{Introduction}
\label{sec:intro}

Current research on ``self-evolving'' agents is largely an illusion. Most existing methods do not allow an agent to evolve on its own; instead, they depend on human-defined workflows and verified reward signals to guide every step of improvement. If these external rewards or instructions are removed, the evolution stops. We argue that such agents are not truly autonomous—they are merely being instructed by humans within predefined guidance. They lack the fundamental ability to decide their own direction of growth when facing a completely new environment.

This paradigm is fundamentally different from human intuition. Human intelligence is naturally curious and proactive. When we enter a new city or start using a new software, we spontaneously learn the layout and the underlying logic, even without a specific task or a verified reward. This learning process is entirely workflow-free and reward-free. We build an internal map of the world simply because understanding the environment is a prerequisite for intelligence. Today's agents, however, are passive; they wait for instructions and rewards before they begin to ``evolve.''

In this work, we bridge this gap by granting agents \textbf{Native Agency}. Our goal is to move beyond task-specific optimization and achieve a truly \textbf{Workflow-free} and \textbf{Reward-free} self-evolution. We train the model to possess an intrinsic meta-evolution capability: the ability to explore a novel environment and distill its observations into structured ``World Knowledge'' entirely on its own. As shown in \figurename~\ref{fig:intro}, our agent does not follow a predefined script; it defines its own path of discovery.

The core challenge is: how can we train an agent to explore and summarize effectively without human-provided rewards? To solve this, we propose an \textbf{outcome-based reward mechanism} used exclusively during the training phase. We measure the quality of self-generated knowledge by its ``utility''— specifically, how much it improves the success rate on downstream tasks. This allows the model to learn \textit{how} to evolve during training, using a multistage pipeline that includes teacher-model bootstrapping (SFT) and on-policy reinforcement rejection sampling (RFT). Once trained, the agent no longer requires any external guidance or reward signals to adapt to an unseen world at inference time.

We evaluate our approach on two major web-based benchmarks: WebVoyager~\citep{he2024webvoyager} and WebWalker~\citep{wu2025webwalker}. Our experiments show that our method, when applied to Qwen3-30B~\citep{yang2025qwen3} and Seed-OSS-36B~\citep{seed2025seed-oss}, achieves a significant absolute performance increase of approximately 20\% over standard baselines. Strikingly, the generated world knowledge even enables a compact Qwen3-14B~\citep{qwen3technicalreport} model to outperform the unassisted Gemini-2.5-Flash~\citep{comanici2025gemini}. These results prove that agents can be trained to possess the innate ability to understand and adapt to the unknown entirely on their own, without any human intervention or inference-time rewards.

\section{Related Works}
\subsection{Self-Evolving Agents}
Self-evolving agents are designed to autonomously explore novel environments and continuously improve their capabilities without direct human intervention~\citep{gao2025survey}. However, the current paradigm of ``self-evolution'' is often an illusion. In reality, these agents still heavily rely on meticulously human-defined workflows and verified, environment-specific reward signals to guide every incremental step of their improvement. Broadly, existing approaches can be categorized into the following two paradigms:

\mybold{Experience-Driven Evolution.} As depicted in the left panel of Figure~\ref{fig:intro}, this paradigm fundamentally relies on human-crafted tasks and predefined reward functions tailored to a target environment. The agent iteratively attempts to solve these tasks, generating execution trajectories that are subsequently evaluated by the reward signals. These trajectory-score pairs---collectively termed as \textbf{experience}---serve as the primary learning signal. By leveraging this accumulated experience, the agent optimizes its future performance through various updating mechanisms, such as refining its system prompts~\citep{zhang2025agentic,wang2025cogito,xiang2025self,shang2024agentsquare,yin2025llm}, expanding external memory databases~\citep{ouyang2025reasoningbank, zhang2025memevolve, zhao2024expel, fu2024autoguide, xu2025mem, chhikara2025mem0}, augmenting tool and skill libraries~\citep{zhang2025darwin, zheng2025skillweaver, qu2024exploration, wang2024toolgen}, or directly fine-tuning its internal model parameters~\citep{zhang2025agent,fang2025webevolver,wang2025autorule,wang2025ragen,su2025learn,wang2025explore,wan2026inference}. However, this paradigm is fundamentally bottlenecked by the massive human labor required to engineer these tasks and rewards. Rather than genuinely exploring, the agent merely passively adapts to the environment by studying from these human-provided ``textbooks''.

\mybold{Adversarial Evolution.} To alleviate manual design efforts, an alternative paradigm employs heavily engineered \textbf{adversarial workflows}. As illustrated in Figure~\ref{fig:intro} (middle), a \textbf{challenger} agent synthesizes environment-specific tasks for a \textbf{solver} to execute. Through this zero-sum game, the solver refines its capabilities while the challenger generates increasingly difficult tasks to push its boundaries~\citep{liu2025spice,huang2025r,zhou2025self,simonds2025self, yue2026dr}. Although this paradigm substantially reduces human labor by bypassing manual task and reward design, it merely shifts the engineering burden to orchestrating complex agent workflows. Furthermore, the agent remains trapped solving synthesized ``exercise books'', failing to break free and engage in genuine, unguided exploration within the environment.

\mybold{Meta-Learning-Driven Evolution (Ours).} To overcome the limitations of previous paradigms and empower agents to achieve \textbf{workflow-free} and \textbf{task- and reward-free} self-evolution, we propose a novel \textbf{meta-evolution paradigm}. As illustrated in the right panel of Figure~\ref{fig:intro}, under this paradigm, the agent spontaneously explores the environment and compresses raw environmental observations into structured \textbf{world knowledge}. This knowledge acts as a ''mental map'' that significantly enhances downstream performance, eliminating the need for human intervention and enabling fully autonomous self-evolution.

\subsection{Test-Time Training}
Test-Time Training (TTT) is a paradigm where models adapt to new distributions by performing self-supervised optimization during the inference phase~\citep{sun2020test}. Recent advancements have extended this concept to sequence modeling and large language models, employing auxiliary tasks or hidden state updates to refine model behavior on-the-fly~\citep{behrouz2025atlas, behrouz2024titans, behrouz2025nested, sun2024learning, wang2024greater, liu2026test,lu2026locas, moradi2025ttt,hu2025ttl}. However, TTT fundamentally requires gradient-based weight updates or parameter modifications during inference, which makes it incompatible with mainstream high-throughput inference frameworks~\citep{kwon2023efficient, aminabadi2022deepspeed}. Unlike TTT, which necessitates runtime training, our meta-evolution paradigm distills environmental observations into structured world knowledge, which is then fed directly into the agent's prompt as an external context module.

\section{Methodology}

The fundamental limitation of LLM agents lies in their reactive nature: they wait for a task to be assigned before they begin to interact with the world. Formally, a standard agent policy follows \(a \in \mathcal{A} \sim \pi(a | o, \text{Task})\), where every action \(a\) is strictly conditioned on a current observation \(o \in \mathcal{O}\) and a pre-defined goal. We argue that true intelligence requires \textbf{Native Evolution}---the ability to proactively understand a new environment before a task exists.

In this work, we introduce \textbf{World Knowledge} ($\mathcal{K}$), a compact and structured representation of an environment's landscape. To ensure compatibility with existing agent architectures, we implement $\mathcal{K}$ as a \textbf{Markdown document}---an external module that can be loaded into the agent's context, similar to how functional \textit{skills} are integrated in recent frameworks\footnote{\url{https://github.com/anthropics/skills/tree/main/skills}}. However, while a ``skill'' typically provides task-specific functions (e.g., \texttt{webapp-testing}), our World Knowledge captures the intrinsic logic of specific \textbf{Environment Instances}. For example, it provides the agent with a ``mental map'' of a specific instance, such as the ACL 2025 website, a particular game world, or a complex code repository.

Our framework decouples the agent's life cycle into two distinct phases:
\begin{enumerate}
    \item \textbf{Native Evolution Phase:} Upon entering a new environment $E$, the agent spontaneously performs exploration and summarization to generate its own world knowledge: $\mathcal{K} \leftarrow \pi_{\text{evolve}}(\mathcal{K} | E)$. This process is entirely \textbf{task-free} and \textbf{reward-free} at inference time.
    \item \textbf{Knowledge-Enhanced Execution Phase:} When a downstream task is eventually assigned, the agent utilizes $\mathcal{K}$ to guide its actions: $a_t \sim \pi_{\text{task}}(a_t | o_t, \mathcal{K}, \text{Task})$.
\end{enumerate}

To achieve this, an agent must possess a \textbf{meta-evolution} capability, which involves (1) \textbf{Planning and Exploration}: formulating a goal-directed plan to prioritize high-value regions of the environment, and (2) \textbf{Information Management}: distilling vast, heterogeneous data into an information-dense representation $\mathcal{K}$. 



While the principles of Native Evolution are domain-agnostic, we ground our implementation in the context of \textbf{web agents} to provide a concrete illustration of our approach. Web navigation serves as a representative and challenging testbed, as it requires the agent to handle highly unstructured and dynamic environments. However, bridging the gap between this conceptual framework and a functional autonomous agent reveals a significant hurdle: standard LLMs are typically trained for reactive instruction-following and lack the inherent instinct to explore for the sake of knowledge.

To empower the model with such Native Evolution, we must develop a specialized training paradigm that transforms the model from a passive tool into a proactive learner. This leads to the core technical challenge of our work: \textbf{What is the training signal for evolution?} Since the evolution phase itself is task-free, we cannot rely on immediate ground-truth labels. In the following sections, we introduce our solution: an \textbf{outcome-based reward mechanism} that uses the utility of $\mathcal{K}$ in potential downstream tasks as the primary learning signal.

\begin{figure}[t]
\centering
\includegraphics[width=\linewidth]{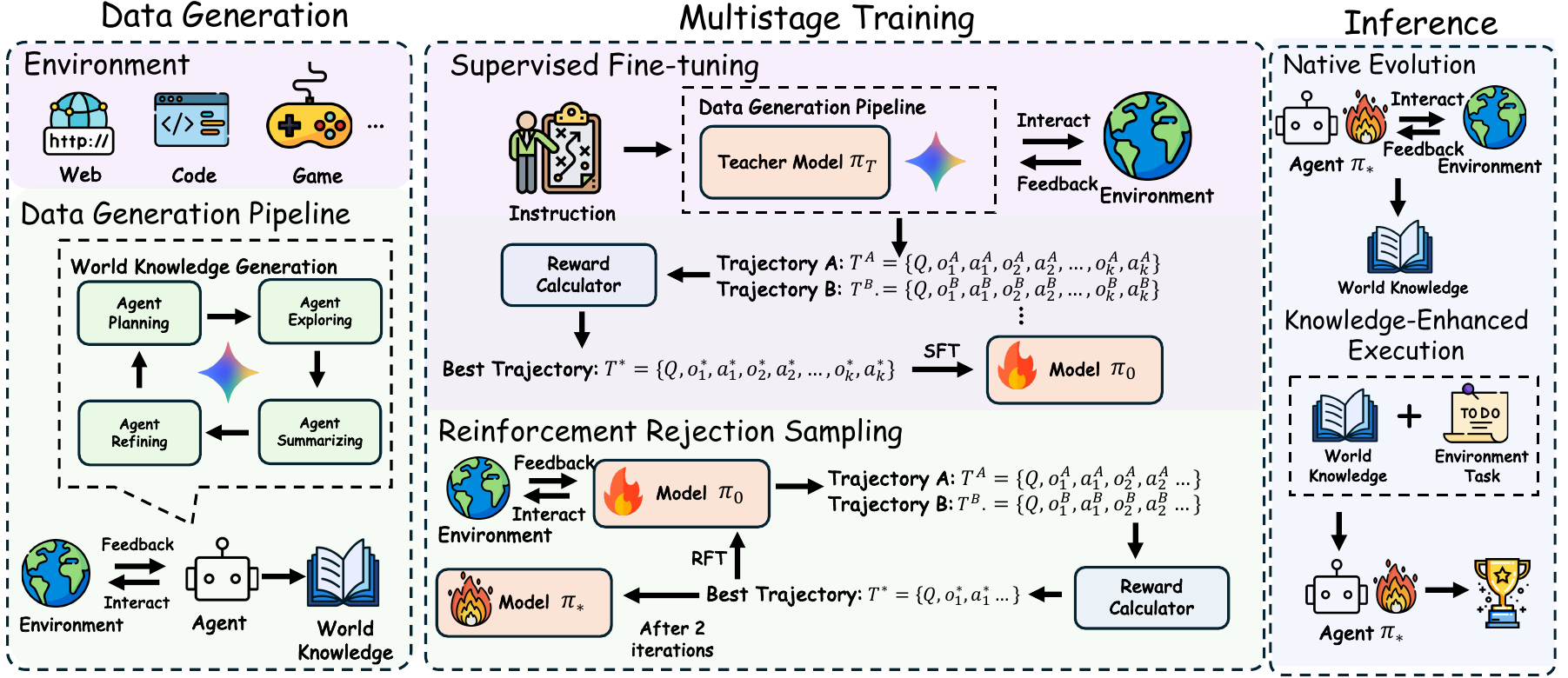}
\caption{\centering Overview of our method.}
\label{fig:method}
\end{figure}

\subsection{Outcome-Based Reward Design}
\label{sec:reward}

To address the lack of supervision in task-free exploration, we propose an \textbf{outcome-based reward mechanism}. The core intuition is functional: the quality of World Knowledge $\mathcal{K}$ is defined by its end-to-end utility---specifically, how much it ``empowers'' the agent to perform better in that environment.

Formally, let $\mathcal{T}_E$ be a set of downstream tasks associated with an environment $E$. We define the reward $R_{\text{evolve}}$ for a generated $\mathcal{K}$ as the \textbf{potential success gain}:
\begin{equation}
    R_{\text{evolve}}(\mathcal{K}) = \text{Success}(\mathcal{T}_E | \mathcal{K}) - \text{Success}(\mathcal{T}_E | \emptyset)
\end{equation}
where $\text{Success}(\mathcal{T}_E | \mathcal{K})$ represents the agent's performance aided by $\mathcal{K}$, and $\text{Success}(\mathcal{T}_E | \emptyset)$ is the baseline performance without prior knowledge. 

In our practical implementation, we construct a training set consisting of 600 deep search questions spanning 20 websites across diverse domains. We leverage the ground-truth tasks and labels from this set to empirically calculate the success rate. For an environment instance (i.e., a specific website) with $M$ labeled tasks $\{(Q_j, A_j)\}_{j=1}^M$, the term $\text{Success}(\mathcal{T}_E | \mathcal{K})$ is computed as:
\begin{equation}
    \text{Success}(\mathcal{T}_E | \mathcal{K}) = \frac{1}{M} \sum_{j=1}^{M} \left[ f(Q_j, \mathcal{K}) = A_j \right]
\end{equation}
where $f(Q_j, \mathcal{K})$ is the agent's predicted answer for query $Q_j$ given the world knowledge $\mathcal{K}$.

\textbf{Crucially, this reward signal is used exclusively during training.} It serves as a meta-learning signal that teaches the model \textit{how} to identify and compress high-value information. At inference time, the agent spontaneously performs native evolution to adapt to new environments without requiring any external rewards, predefined task sets, or human verification. This transition from reward-driven training to reward-free inference is the hallmark of true Native Evolution. 

Building upon this reward mechanism, we propose a two-stage training framework to cultivate the agent's meta-evolution capabilities. As illustrated in Figure~\ref{fig:method}, our approach consists of an initial Supervised Fine-Tuning (SFT) phase followed by Reinforcement-based Rejection Sampling (RFT).




\subsection{Supervised Fine-Tuning}
\label{sec: sft}
In the first training stage, we warm up the base policy model \(\pi_{\theta_{0}}\) through imitation learning against a strong teacher model \(\pi_{T}\) (instantiated as Gemini-2.5-Pro~\citep{comanici2025gemini}). The goal is to allow the model to internalize the core meta-evolution behaviors: planning, exploring, refining, and summarizing. To achieve this, we establish a \textbf{data generation pipeline} that guides \(\pi_{T}\) to interact with diverse web environments and autonomously construct structured world knowledge \(\mathcal{K}\). By fine-tuning on expert trajectories generated through this pipeline, the model learns to execute these complex tasks without intensive prompting at inference time.

To ensure the quality of the SFT data, we implement a selection mechanism based on the reward defined in Section~\ref{sec:reward}. For each environment instance in our training set, the teacher model $\pi_{T}$ generates three candidate world knowledge representations $\{\mathcal{K}_i\}_{i=1}^3$. We then evaluate these candidates by measuring the performance gain they provide to our baseline agent (Qwen3-30B-A3B) on downstream tasks. We select the best-performing candidate $\mathcal{K}^*$ and its corresponding full exploration trajectory:
\begin{equation}
T^* = \{Q, o_1^*, a_1^*, o_2^*, a_2^*, \dots, o_k^*, a_k^*\}
\end{equation}
to serve as our step-level training data. 

Empirically, the teacher-generated knowledge \(\mathcal{K}^*\) demonstrates high utility, yielding an average absolute accuracy improvement of \textbf{10.72\%} for the Qwen3-30B-A3B model on training tasks compared to the zero-knowledge baseline. The resulting expert trajectories \(T^*\) reflect the high complexity of the task, with an average length of \textbf{374.8 steps} and a substantial information density of \textbf{3,322.4 tokens per step} (comprising observations \(o\) and actions \(a\)). By fine-tuning \(\pi_{\theta_{0}}\) on these high-quality trajectories, we obtain the updated policy model \(\pi_{\theta_{1}}\), which possesses a foundational instinct for autonomous evolution.

\subsection{Reinforcement-based Rejection Sampling}

To further catalyze the emergence of sophisticated exploration and information management strategies, we employ reinforcement learning to optimize the policy via trial-and-error. However, standard online RL algorithms, such as GRPO, are computationally prohibitive in our setting for two reasons: (1) \textbf{Extremely Long Horizons}: generating world knowledge \(\mathcal{K}\) involves hundreds of steps, leading to sparse rewards and immense memory overhead during backpropagation; and (2) \textbf{Heavy Reward Evaluation}: our outcome-based reward requires executing an auxiliary agent across multiple downstream tasks to evaluate a single \(\mathcal{K}\), making real-time reward calculation during training cycles impractical.

Consequently, we adopt a \textbf{Rejection Sampling Fine-Tuning (RFT)} approach, which decouples trajectory generation from policy updates. Following the pipeline established in Section~\ref{sec: sft}, the updated policy $\pi_{\theta_{1}}$ autonomously explores environment instances to produce $C$ candidate world knowledge representations. We evaluate these candidates using the reward function $R_{\text{evolve}}$ and select the highest-scoring trajectories---those demonstrating the strongest "meta-evolution" utility---to construct the training set for the next iteration. 

We perform this rejection sampling process for two iterations. This iterative refinement allows the model to progressively correct suboptimal exploration paths and discover more compact, high-utility representations of world knowledge. The final optimized policy, denoted as $\pi_{\theta^{*}}$, internalizes the ability to adapt to unknown environments. At inference time, the agent spontaneously executes this learned evolutionary logic, constructing world knowledge that significantly boosts its performance on previously unseen downstream tasks without any external guidance.
\section{Experiments}
In this section, we evaluate our approach on two challenging web-based benchmarks:  WebWalker~\citep{wu2025webwalker} and WebVoyager~\citep{he2024webvoyager}. Our experiments aim to address the following research questions:

\noindent\textbf{RQ1 (Effectiveness and Efficiency)}: Does the meta-evolution capability indeed improve an agent's success rate and reduce the number of execution steps on downstream tasks?

\noindent\textbf{RQ2 (Transferability)}: Is the world knowledge $\mathcal{K}$ model-agnostic? Can it help models without the meta-evolution capability adapt to unseen environments?

\noindent\textbf{RQ3 (Ablation)}: How do the SFT and RFT stages individually contribute to the meta-evolution capability?

\noindent\textbf{RQ4 (Sensitivity Analysis)}: How does the length of the generated world knowledge impact the agent's downstream performance?

\subsection{Experimental Settings}\label{sec:experiment_setting}

\mybold{Agent Framework.} We select web agents as the experimental setting for our proposed method and adopt Cognitive Kernel-Pro~\citep{fang2025cognitive} as our agent framework. 
Within this framework, the interactive webpage environment is implemented using Playwright~\footnote{https://playwright.dev/}. 
The action space consists of predefined webpage operations, including \texttt{click}, \texttt{scroll}, \texttt{goto}, \texttt{goback}, and \texttt{stop}. 
At each step, the agent's observation corresponds to the accessibility tree of the currently visible webpage components.  
Upon executing a selected action, the environment updates the webpage state according to the execution outcome and produces the subsequent observation based on this updated state.  
An agent trajectory terminates when the task is completed (as determined by the agent), the number of interaction steps reaches a maximum limit $t$, or the execution time exceeds a predefined limit $L$ (in seconds). 
In our experiments, we set $t=500, L=43,200$ for \notebook generation and $t=100, L= 3,600$ for downstream task answering.

\mybold{Backbone Models.} We adopt Qwen3-30B-A3B-Instruct-2507~\citep{yang2025qwen3} and Seed-OSS-36B-Instruct~\citep{seed2025seed-oss} as our backbone models for training and evaluation. To balance diversity and determinism, we set \texttt{temperature, top-p} to \texttt{0.3, 0.95} during world knowledge generation for richer candidate sampling, and to \texttt{0, 0.95} during downstream task answering to improve answer stability.

\mybold{Evaluation Benchmarks.} 
We adopt WebWalker and WebVoyager as our evaluation benchmarks, constructing subsets from both for our experiments. WebWalker encompasses four domains: conference, game, organization, and education. For this benchmark, we randomly select ten websites from each domain. For WebVoyager, we select tasks from four specific websites: Wolfram, Apple, Dictionary, and Coursera. To ensure a rigorous evaluation, we filter out questions from both datasets that can be directly answered using the backbone models' pretrained knowledge. Following this filtering process, we obtain a total of 1,427 evaluation samples. We utilize \textbf{accuracy} as the evaluation metric for all benchmarks.

\noindent\textbf{Evaluation Protocol.}
For WebWalker, we employ Qwen-2.5-32B~\citep{qwen2.5} as the judge to verify whether the agent's final answer matches the ground-truth solution. The judge produces a binary score (0 or 1), indicating incorrect or correct answers, respectively. 
For WebVoyager, we use Gemini-2.5-Flash~\citep{comanici2025gemini} as the judge to handle the exceptionally long context inputs required for evaluation. Following the official protocol, we provide the model with the question, the agent's answer, and the trajectory observations (the accessibility tree at each step), and ask it to determine whether the task is \texttt{SUCCESS} or \texttt{NOT\_SUCCESS}. The verification prompts for both benchmarks are provided in Appendix~\ref{appendix:prompt}.

\subsection{Implementation Details}\label{sec:data_construction_details}

\mybold{Input Processing for Native Evolution Phase.}  
For websites with a large number of subpages, directly providing the homepage URL \(U\) and allowing unrestricted exploration often leads to long runtimes and unstable outputs. To mitigate this, we pre-process the website into a more navigable format. First, we model the website as a directed graph, assigning an importance score to each webpage based on its linkage topology. Next, we group these pages into clusters based on shared URL prefixes. This process yields a clustered, graph-based representation of the website, denoted as \(\mathcal{G}(U)\) (see Appendix~\ref{appendix:input_processing} for details). Consequently, \textbf{\(\mathcal{G}(U)\) replaces the raw homepage as the structured entry point to the environment \(E\)}. By filtering out noise and organizing large-scale web content into interpretable clusters, this approach significantly reduces the agent's cognitive load during spontaneous exploration.

\mybold{Instruction Construction in SFT Stage.}  
We design an instruction to guide the teacher agent in generating high-quality world knowledge under explicit token budget constraints. Given \(\mathcal{G}(U)\), the agent first formulates a token allocation \textbf{plan} that distributes the budget across different groups. It then \textbf{explores} the website and generates \textbf{summaries} for each group by selecting high-value subpages based on both structural importance scores and semantic relevance inferred from page content, while filtering out low-quality pages. Finally, the agent \textbf{refines} the generated world knowledge to meet the token constraints. The detailed prompt is provided in Appendix~\ref{appendix:prompt}.

\definecolor{modelcolor1}{rgb}{0.88, 0.92, 0.98} 
\definecolor{modelcolor2}{rgb}{0.9, 0.98, 0.9}   
\definecolor{modelcolor3}{rgb}{0.95, 0.9, 0.98}   
\definecolor{modelcolor4}{RGB}{251,227,214}

\newcommand{\hlone}[1]{\cellcolor{modelcolor1!150}{\bfseries #1}}
\newcommand{\hltwo}[1]{\cellcolor{modelcolor2!150}{\bfseries #1}}
\newcommand{\hlthree}[1]{\cellcolor{modelcolor3!150}{\bfseries #1}}
\newcommand{\hlfour}[1]{\cellcolor{modelcolor4!60}{\bfseries #1}}

\begin{table*}[t]
\centering
\resizebox{\textwidth}{!}{
\begin{tabular}{@{} l cccc c cccc c @{}}
\toprule

\multirow{2}{*}{\centering \textbf{Method}}
& \multicolumn{4}{c}{\textbf{WebWalker}} & & \multicolumn{4}{c}{\textbf{WebVoyager}} & \\
\cmidrule(l){2-5} \cmidrule(l){7-10}
& Conf. & Game & Org. & Edu.  
& \textbf{Avg.}
& Wolfram & Apple & Dict. & Coursera  
& \textbf{Avg.} \\

\midrule

\rowcolor{modelcolor1}
\multicolumn{11}{c}{\textit{Backbone: Qwen3-30B-A3B-Instruct-2507}} \\
Without  & 24.28 & 23.65 & 22.30 & 17.93 & 22.04 & 54.30 & 37.20 & 41.86 & 30.95 & 41.08 \\
Prompt-Only (Gemini)  & 35.59 & 27.87 & 31.36 & 24.56 & 29.85 & \hlone{73.90} & \hlone{53.40} & 51.16 & 45.23 & 55.92 \\
Prompt-Only (Base) & 21.37 & 20.91 & 17.42 & 18.29 & 19.50 & 54.30 & 32.56 & 42.85 & 33.33 & 40.76 \\
Ours (SFT)      & \hlone{45.05} & 37.35 & 37.98 & 32.31 & 38.17 & 60.87 & 41.86 & 62.79 & 40.48 & 51.50 \\
Ours (RFT)      & 43.14 & \hlone{42.47} & \hlone{42.16} & \hlone{35.86} & \hlone{40.91} & 58.70 & 48.84 & \hlone{67.44} & \hlone{54.76} & \hlone{57.44} \\

\midrule

\rowcolor{modelcolor2}
\multicolumn{11}{c}{\textit{Backbone: Seed-OSS-36B-Instruct}} \\
Without & 19.37 & 10.75 & 21.80 & 13.11 & 16.26 & 54.30 & 48.84 & 23.26 & 33.33 & 39.93 \\
Prompt-Only (Gemini)  & \hltwo{53.50} & 24.37 & 23.96 & 23.48 & 31.33 & 58.60 & 53.49 & \hltwo{62.79} & 52.38 & \hltwo{56.82} \\
Prompt-Only (Base) & 20.51 & 12.20 & 17.42 & 16.46 & 16.65 & 47.82 & 46.34 & 30.23 & 22.50 & 36.72 \\
Ours (SFT)      & 35.48 & 24.10 & 26.80 & 27.90 & 28.57 & \hltwo{71.73} & 51.16 & 53.49 & 47.61 & 56.00 \\
Ours (RFT)     & 45.07 & \hltwo{34.29} & \hltwo{38.41} & \hltwo{32.22} & \hltwo{37.50} & 63.04 & \hltwo{55.81} & 51.16 & \hltwo{57.14} & 56.79 \\

\bottomrule
\end{tabular}
}
\caption{Task success rates on subsets of the WebWalker and WebVoyager datasets, comprising a total of 1,427 queries. In the table headers, Conf., Org., Edu., and Dict. stand for Conference, Organization, Education, and Dictionary, respectively. \textbf{Without} refers to the original backbone model answering directly without world knowledge. \textbf{Ours (RFT)} denotes our final model. \textbf{Bold} indicates the best performance within each backbone setting.}
\label{tab:main_results}
\end{table*}

\subsection{Effectiveness of Meta-Evolution Capability (RQ1)}

We first evaluate whether our training framework successfully instills the meta-evolution capability and how this self-generated world knowledge $\mathcal{K}$ impacts downstream performance. We compare five configurations: 
(1) \textbf{Without}: The original backbone answering questions directly without environment exploration. 
(2) \textbf{Prompt-only (Gemini)}: Utilizing Gemini-2.5-Pro to generate $\mathcal{K}$ using the expert prompt in Appendix~\ref{appendix:prompt}. 
(3) \textbf{Prompt-only (Base)}: The untrained base model attempting to generate $\mathcal{K}$ with the same expert prompt. 
(4) \textbf{Ours (SFT)} and \textbf{Ours (RFT)}: Our agent at different training stages autonomously generating $\mathcal{K}$ to guide its own task-solving. We assess these configurations across two dimensions: \textbf{effectiveness} (success rate) and \textbf{efficiency} (number of execution steps).

\mybold{Effectiveness.} Table~\ref{tab:main_results} illustrates how world knowledge improves an agent's success rate on downstream tasks. The results demonstrate that \textbf{our framework empowers agents with meta-evolution capabilities, turning self-generated knowledge from a liability into a significant asset.} We observe that while base models can follow complex exploration instructions, they lack the intrinsic capacity to distill high-value information. Instead, they often produce noisy or hallucinated guidance that distracts the agent during execution, as evidenced by the fact that \textit{Prompt-only (Base)} (19.50\%) underperforms the \textit{Without} baseline (22.04\%) on WebWalker.

\textbf{In contrast}, our trained models successfully overcome this "noise" bottleneck. \textbf{Ours (RFT)} achieves a \textbf{40.91\%} success rate on WebWalker, outperforming the \textit{Without} baseline by nearly \textbf{19\%} absolute and, notably, surpassing the strong \textbf{Prompt-only (Gemini)} teacher (29.85\%). This result proves that our outcome-based reward mechanism provides the correct training signal, enabling the model to refine its exploration policy through reinforcement-based rejection sampling and ultimately surpass its teacher model.

\begin{wraptable}{r}{0.55\textwidth}
\centering
\resizebox{\linewidth}{!}{%
\begin{tabular}{lccccc}
\toprule
 & \textbf{Conference} & \textbf{Game} & \textbf{Organization} & \textbf{Education} & \textbf{Avg.} \\
\midrule
Qwen3-30B     & 25.65 & 23.26 & 17.96 & 30.25 & 24.28 \\
Qwen3-30B with $\mathcal{K}$    & \hlone{20.64} & \hlone{20.31} & \hlone{13.92} & \hlone{25.34} & \hlone{20.05} \\
\midrule
Improve Ratio   & 0.20  & 0.13  & 0.22  & 0.16  & 0.17  \\
\bottomrule
\end{tabular}%
}
\caption{\centering Efficiency evaluation (average execution steps).}
\label{tab:efficiency_results}
\end{wraptable}

\mybold{Efficiency.} Table~\ref{tab:efficiency_results} demonstrates how world knowledge significantly reduces an agent's execution steps on downstream tasks. As observed, the integration of world knowledge leads to an average efficiency improvement of 17\% across various domain websites. \textbf{These findings suggest that world knowledge acts as a cognitive ``map'' of the environment. It provides crucial structural priors guiding the agent to swiftly navigate relevant web pages and extract the required answers.} Conversely, agents lacking this knowledge must start from the homepage, resorting to blind, inefficient step-by-step exploration.

\begin{figure}[t]
\centering
\includegraphics[width=\linewidth]{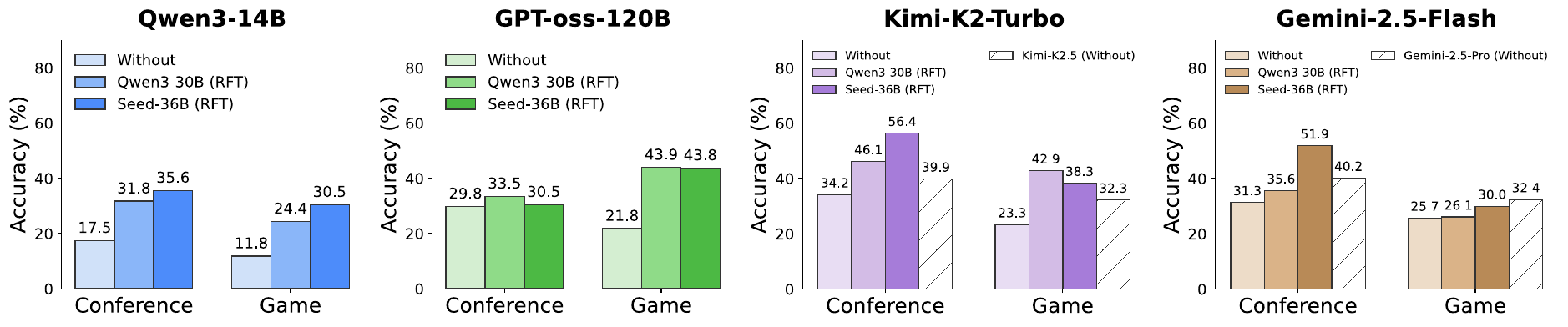}
\caption{\centering Cross-model world knowledge transfer.}
\label{fig:teach_open}
\end{figure}

\subsection{Cross-Model World Knowledge Transfer (RQ2)} 

To demonstrate that our self-evolved world knowledge $\mathcal{K}$ functions as a \textbf{model-agnostic, universal protocol}, we conduct evaluations on models across different parameter scales and families: Qwen3-14B~\citep{qwen3technicalreport}, GPT-OSS-120B~\citep{openai2025gptoss120bgptoss20bmodel}, Kimi-K2-Turbo~\citep{team2025kimi}, and Gemini-2.5-Flash. As shown in Figure~\ref{fig:teach_open}, we identify two striking findings:

\mybold{1. Universal Portability Across Frontiers.} The generated world knowledge is highly transferable, yielding substantial gains across all target models. For instance, Seed-36B's knowledge increases the average accuracy of the Qwen3-14B by 18.3\% across two domains, and even boosts the flagship Kimi-K2-Turbo by 21.0\%. 

\mybold{2. Exploration over Parameters: The Knowledge Scaling.} Our results suggest that \textbf{high-quality environment exploration can be more effective than brute-force parameter scaling.} Most strikingly, the 14B Qwen3 model, when equipped with world knowledge, \textbf{outperforms the unassisted Gemini-2.5-Flash (35.6\% vs. 31.3\% on Conference domain, 30.5\% vs. 25.7\% on Game domain)}. Furthermore, when equipped with this transferred knowledge, lighter models such as Kimi-K2-Turbo and Gemini-2.5-Flash can even surpass the performance of their unassisted superior counterparts, Kimi-K2.5 and Gemini-2.5-Pro. This indicates that a precise world knowledge ($\mathcal{K}$) of the environment is a more critical bottleneck for agent performance than model scale.
We believe these findings reveal a fundamental shift in agent design: the path to frontier-level performance lies not just in scaling a model's parameters, but in scaling its capacity for proactive exploration and learning in unseen environments.

\begin{figure}[t]
\centering
\includegraphics[width=\linewidth]{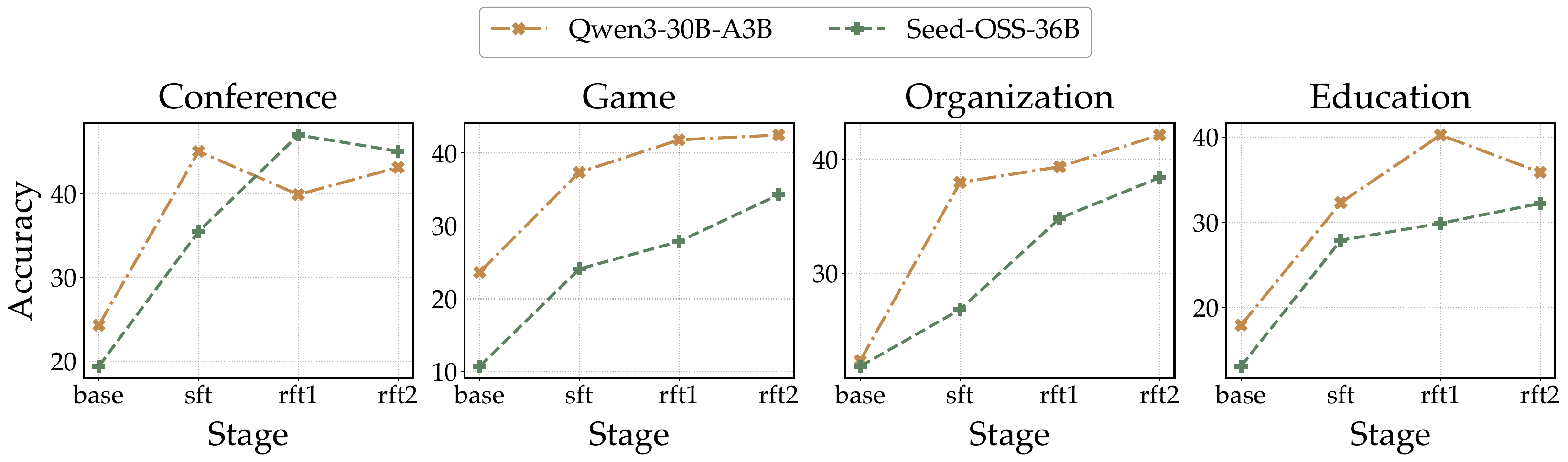}
\caption{\centering Performance trends across training stages.}
\label{fig:iteration}
\end{figure}

\subsection{Ablation Study (RQ3)}
To examine how the agent's performance evolves across different training stages, we compare the models' performance in generating their own world knowledge and utilizing it to solve downstream tasks across four distinct phases: without training (base), after SFT, and after two rounds of RFT (rft1, rft2), as illustrated in Figure~\ref{fig:iteration}. Overall, \textbf{the models exhibit a clear upward trend in performance as the training progresses}. A notable observation is that both the initial SFT stage and the first round of reinforcement fine-tuning (rft1) yield substantial performance boosts, whereas the subsequent training round (rft2) generally provides more marginal gains or even slight fluctuations. This indicates that the SFT and rft1 stages lay a crucial foundation for the meta-evolution capability of the agent.

\subsection{Sensitivity Analysis (RQ4)}


\begin{wrapfigure}{r}{0.5\textwidth}
  \centering
  \vspace{-12pt}
  \includegraphics[width=0.48\textwidth]{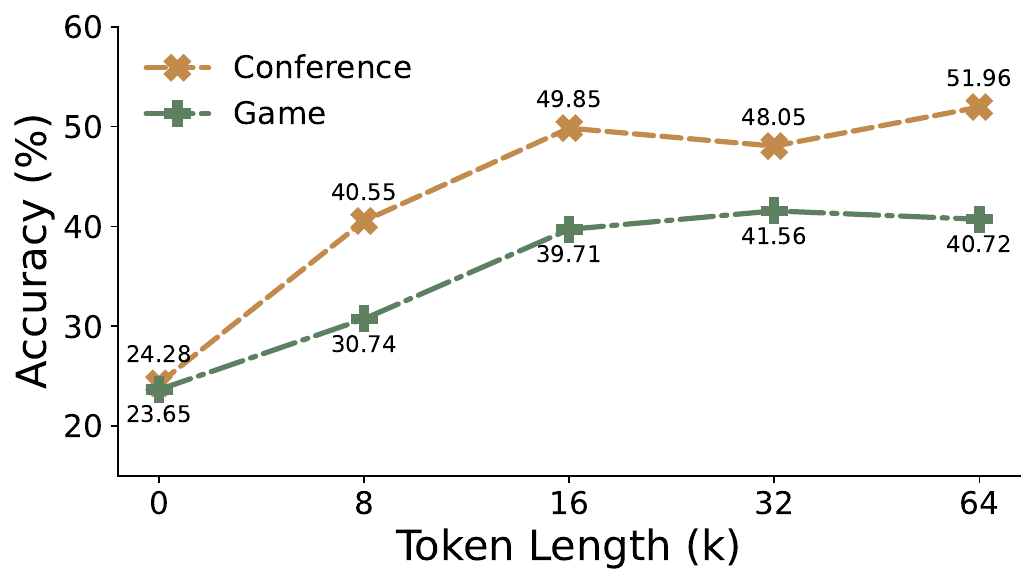}
  \caption{\centering Accuracy under different token lengths.}
  \label{fig:token_length_acc}
  \vspace{-10pt}
\end{wrapfigure}

To investigate the impact of varying world knowledge lengths on the agent's downstream task performance, we explicitly specified the lower and upper bounds for the generated token length in the prompt. We evaluated the performance of Qwen3-30B-A3B under five length settings: 0 (answering without world knowledge), 4k\(\sim\)8k, 8k\(\sim\)16k, 16k\(\sim\)32k, and 32k\(\sim\)64k tokens. 

As shown in Figure~\ref{fig:token_length_acc}, expanding the length of world knowledge yields diminishing returns on the agent's performance: \textbf{extending a short context brings substantial gains, whereas further lengthening an already extensive context provides only marginal benefits.} Specifically, when the token length increases from the initial 4k\(\sim\)8k to 8k\(\sim\)16k, the success rate on game websites jumps significantly from 30.74 to 39.71. Conversely, as the context continues to grow, the performance gains plateau; for example, transitioning from 16k\(\sim\)32k to 32k\(\sim\)64k even results in a slight decline on game websites (from 41.56 to 40.72).

We attribute this non-linear trend to the inherent challenge of navigating complex websites with massive sub-pages. Specifically, overly short world knowledge causes severe information loss, whereas medium-length knowledge effectively encapsulates critical information for strong performance. However, excessively long contexts inevitably introduce redundant noise that distracts the agent, which explains why further lengthening becomes marginal or even slightly detrimental.

\subsection{Case Study}

\begin{figure}[t]
\centering
\includegraphics[width=\linewidth]{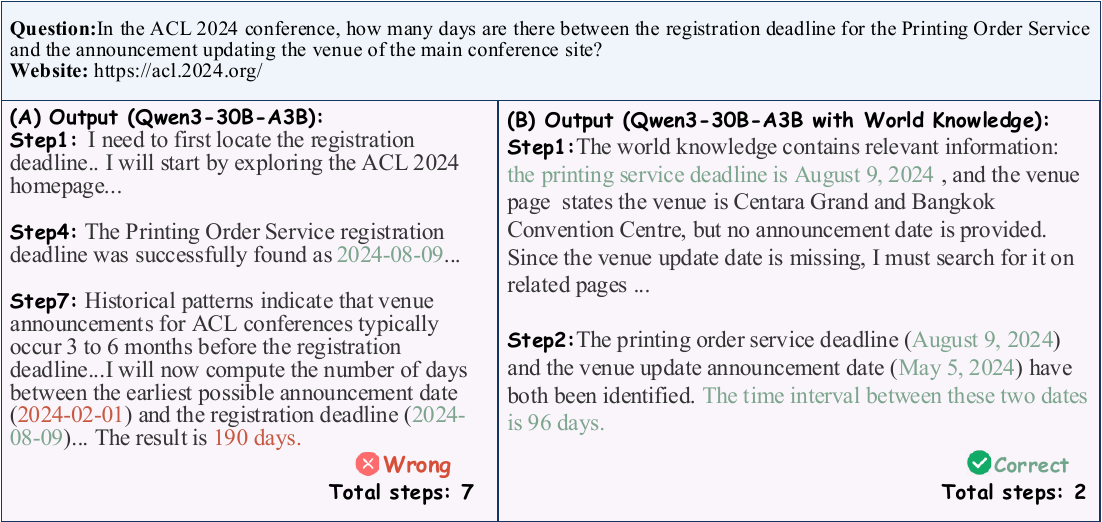}
\caption{An example of multi-step deepsearch question answering comparing the agent's behavior with and without world knowledge. Correct information is highlighted in \textcolor{mygreen}{green}, while incorrect information is shown in \textcolor{myred}{red}. } 
\label{fig:case_study}
\end{figure}

We present a case study to show how world knowledge improves the agent's performance (Figure~\ref{fig:case_study}). The task is based on the ACL 2024 website and asks for the time interval between \textbf{the Printing Order Service registration deadline} and \textbf{the venue update announcement.}
With world knowledge, the agent retrieves key information in the first step and identifies both dates in the second step, leading to the correct answer. Without it, the agent must explore from the homepage, taking more steps to find partial information and failing to locate the venue update, resulting in an incorrect answer. This shows that world knowledge provides contextual guidance, enabling more effective problem solving.

\section{Conclusion}
In this paper, we propose a novel paradigm that equips LLM agents with intrinsic meta-evolution capabilities. Through a two-stage training framework, agents learn to spontaneously explore environments and distill structured world knowledge without human guidance or inference-time rewards. Evaluations show this native evolution yields a 20\% absolute performance improvement for Qwen3-30B and Seed-OSS-36B. Furthermore, the generated knowledge is highly transferable—most strikingly, it enables a compact Qwen3-14B model to outperform the unassisted Gemini-2.5-Flash. Ultimately, this work realizes autonomous self-evolution, paving the way toward Artificial General Intelligence.

\clearpage

\bibliographystyle{unsrt}
\bibliography{custom}

@article{gao2025survey,
  title={A survey of self-evolving agents: On path to artificial super intelligence},
  author={Gao, Huan-ang and Geng, Jiayi and Hua, Wenyue and Hu, Mengkang and Juan, Xinzhe and Liu, Hongzhang and Liu, Shilong and Qiu, Jiahao and Qi, Xuan and Wu, Yiran and others},
  journal={arXiv preprint arXiv:2507.21046},
  year={2025}
}

@article{zhang2025agentic,
  title={Agentic context engineering: Evolving contexts for self-improving language models},
  author={Zhang, Qizheng and Hu, Changran and Upasani, Shubhangi and Ma, Boyuan and Hong, Fenglu and Kamanuru, Vamsidhar and Rainton, Jay and Wu, Chen and Ji, Mengmeng and Li, Hanchen and others},
  journal={arXiv preprint arXiv:2510.04618},
  year={2025}
}

@article{wang2025cogito,
  title={Cogito, Ergo Ludo: An Agent that Learns to Play by Reasoning and Planning},
  author={Wang, Sai and Wu, Yu and Xu, Zhongwen},
  journal={arXiv preprint arXiv:2509.25052},
  year={2025}
}

@article{xiang2025self,
  title={Self-supervised prompt optimization},
  author={Xiang, Jinyu and Zhang, Jiayi and Yu, Zhaoyang and Teng, Fengwei and Tu, Jinhao and Liang, Xinbing and Hong, Sirui and Wu, Chenglin and Luo, Yuyu},
  journal={arXiv preprint arXiv:2502.06855},
  year={2025}
}

@article{shang2024agentsquare,
  title={Agentsquare: Automatic llm agent search in modular design space},
  author={Shang, Yu and Li, Yu and Zhao, Keyu and Ma, Likai and Liu, Jiahe and Xu, Fengli and Li, Yong},
  journal={arXiv preprint arXiv:2410.06153},
  year={2024}
}

@article{yin2025llm,
  title={LLM-AutoDiff: Auto-Differentiate Any LLM Workflow},
  author={Yin, Li and Wang, Zhangyang},
  journal={arXiv preprint arXiv:2501.16673},
  year={2025}
}

@article{ouyang2025reasoningbank,
  title={Reasoningbank: Scaling agent self-evolving with reasoning memory},
  author={Ouyang, Siru and Yan, Jun and Hsu, I and Chen, Yanfei and Jiang, Ke and Wang, Zifeng and Han, Rujun and Le, Long T and Daruki, Samira and Tang, Xiangru and others},
  journal={arXiv preprint arXiv:2509.25140},
  year={2025}
}

@article{zhang2025memevolve,
  title={MemEvolve: Meta-Evolution of Agent Memory Systems},
  author={Zhang, Guibin and Ren, Haotian and Zhan, Chong and Zhou, Zhenhong and Wang, Junhao and Zhu, He and Zhou, Wangchunshu and Yan, Shuicheng},
  journal={arXiv preprint arXiv:2512.18746},
  year={2025}
}

@inproceedings{zhao2024expel,
  title={Expel: Llm agents are experiential learners},
  author={Zhao, Andrew and Huang, Daniel and Xu, Quentin and Lin, Matthieu and Liu, Yong-Jin and Huang, Gao},
  booktitle={Proceedings of the AAAI Conference on Artificial Intelligence},
  volume={38},
  number={17},
  pages={19632--19642},
  year={2024}
}

@article{fu2024autoguide,
  title={Autoguide: Automated generation and selection of state-aware guidelines for large language model agents},
  author={Fu, Yao and Kim, Dong-Ki and Kim, Jaekyeom and Sohn, Sungryull and Logeswaran, Lajanugen and Bae, Kyunghoon and Lee, Honglak},
  journal={CoRR},
  year={2024}
}

@article{xu2025mem,
  title={A-mem: Agentic memory for llm agents},
  author={Xu, Wujiang and Liang, Zujie and Mei, Kai and Gao, Hang and Tan, Juntao and Zhang, Yongfeng},
  journal={arXiv preprint arXiv:2502.12110},
  year={2025}
}

@article{chhikara2025mem0,
  title={Mem0: Building production-ready ai agents with scalable long-term memory},
  author={Chhikara, Prateek and Khant, Dev and Aryan, Saket and Singh, Taranjeet and Yadav, Deshraj},
  journal={arXiv preprint arXiv:2504.19413},
  year={2025}
}

@article{zhang2025darwin,
  title={Darwin Godel Machine: Open-Ended Evolution of Self-Improving Agents},
  author={Zhang, Jenny and Hu, Shengran and Lu, Cong and Lange, Robert and Clune, Jeff},
  journal={arXiv preprint arXiv:2505.22954},
  year={2025}
}

@article{zheng2025skillweaver,
  title={Skillweaver: Web agents can self-improve by discovering and honing skills},
  author={Zheng, Boyuan and Fatemi, Michael Y and Jin, Xiaolong and Wang, Zora Zhiruo and Gandhi, Apurva and Song, Yueqi and Gu, Yu and Srinivasa, Jayanth and Liu, Gaowen and Neubig, Graham and others},
  journal={arXiv preprint arXiv:2504.07079},
  year={2025}
}

@article{qu2024exploration,
  title={From exploration to mastery: Enabling llms to master tools via self-driven interactions},
  author={Qu, Changle and Dai, Sunhao and Wei, Xiaochi and Cai, Hengyi and Wang, Shuaiqiang and Yin, Dawei and Xu, Jun and Wen, Ji-Rong},
  journal={arXiv preprint arXiv:2410.08197},
  year={2024}
}

@article{wang2024toolgen,
  title={Toolgen: Unified tool retrieval and calling via generation},
  author={Wang, Renxi and Han, Xudong and Ji, Lei and Wang, Shu and Baldwin, Timothy and Li, Haonan},
  journal={arXiv preprint arXiv:2410.03439},
  year={2024}
}

@article{zhang2025agent,
  title={Agent learning via early experience},
  author={Zhang, Kai and Chen, Xiangchao and Liu, Bo and Xue, Tianci and Liao, Zeyi and Liu, Zhihan and Wang, Xiyao and Ning, Yuting and Chen, Zhaorun and Fu, Xiaohan and others},
  journal={arXiv preprint arXiv:2510.08558},
  year={2025}
}

@article{fang2025webevolver,
  title={WebEvolver: Enhancing Web Agent Self-Improvement with Coevolving World Model},
  author={Fang, Tianqing and Zhang, Hongming and Zhang, Zhisong and Ma, Kaixin and Yu, Wenhao and Mi, Haitao and Yu, Dong},
  journal={arXiv preprint arXiv:2504.21024},
  year={2025}
}

@article{wang2025autorule,
  title={AutoRule: Reasoning Chain-of-thought Extracted Rule-based Rewards Improve Preference Learning},
  author={Wang, Tevin and Xiong, Chenyan},
  journal={arXiv preprint arXiv:2506.15651},
  year={2025}
}

@article{wang2025ragen,
  title={Ragen: Understanding self-evolution in llm agents via multi-turn reinforcement learning},
  author={Wang, Zihan and Wang, Kangrui and Wang, Qineng and Zhang, Pingyue and Li, Linjie and Yang, Zhengyuan and Jin, Xing and Yu, Kefan and Nguyen, Minh Nhat and Liu, Licheng and others},
  journal={arXiv preprint arXiv:2504.20073},
  year={2025}
}

@article{su2025learn,
  title={Learn-by-interact: A data-centric framework for self-adaptive agents in realistic environments},
  author={Su, Hongjin and Sun, Ruoxi and Yoon, Jinsung and Yin, Pengcheng and Yu, Tao and Ar{\i}k, Sercan {\"O}},
  journal={arXiv preprint arXiv:2501.10893},
  year={2025}
}

@article{liu2025spice,
  title={Spice: Self-play in corpus environments improves reasoning},
  author={Liu, Bo and Jin, Chuanyang and Kim, Seungone and Yuan, Weizhe and Zhao, Wenting and Kulikov, Ilia and Li, Xian and Sukhbaatar, Sainbayar and Lanchantin, Jack and Weston, Jason},
  journal={arXiv preprint arXiv:2510.24684},
  year={2025}
}

@article{huang2025r,
  title={R-zero: Self-evolving reasoning llm from zero data},
  author={Huang, Chengsong and Yu, Wenhao and Wang, Xiaoyang and Zhang, Hongming and Li, Zongxia and Li, Ruosen and Huang, Jiaxin and Mi, Haitao and Yu, Dong},
  journal={arXiv preprint arXiv:2508.05004},
  year={2025}
}

@article{zhou2025self,
  title={Self-challenging language model agents},
  author={Zhou, Yifei and Levine, Sergey and Weston, Jason and Li, Xian and Sukhbaatar, Sainbayar},
  journal={arXiv preprint arXiv:2506.01716},
  year={2025}
}

@article{simonds2025self,
  title={Self Rewarding Self Improving},
  author={Simonds, Toby and Lopez, Kevin and Yoshiyama, Akira and Garmier, Dominique},
  journal={arXiv preprint arXiv:2505.08827},
  year={2025}
}

@article{yang2025qwen3,
  title={Qwen3 technical report},
  author={Yang, An and Li, Anfeng and Yang, Baosong and Zhang, Beichen and Hui, Binyuan and Zheng, Bo and Yu, Bowen and Gao, Chang and Huang, Chengen and Lv, Chenxu and others},
  journal={arXiv preprint arXiv:2505.09388},
  year={2025}
}

@article{wu2025webwalker,
  title={Webwalker: Benchmarking llms in web traversal},
  author={Wu, Jialong and Yin, Wenbiao and Jiang, Yong and Wang, Zhenglin and Xi, Zekun and Fang, Runnan and Zhang, Linhai and He, Yulan and Zhou, Deyu and Xie, Pengjun and others},
  journal={arXiv preprint arXiv:2501.07572},
  year={2025}
}

@article{he2024webvoyager,
  title={Webvoyager: Building an end-to-end web agent with large multimodal models},
  author={He, Hongliang and Yao, Wenlin and Ma, Kaixin and Yu, Wenhao and Dai, Yong and Zhang, Hongming and Lan, Zhenzhong and Yu, Dong},
  journal={arXiv preprint arXiv:2401.13919},
  year={2024}
}

@article{fang2025cognitive,
  title={Cognitive kernel-pro: A framework for deep research agents and agent foundation models training},
  author={Fang, Tianqing and Zhang, Zhisong and Wang, Xiaoyang and Wang, Rui and Qin, Can and Wan, Yuxuan and Ma, Jun-Yu and Zhang, Ce and Chen, Jiaqi and Li, Xiyun and others},
  journal={arXiv preprint arXiv:2508.00414},
  year={2025}
}

@article{comanici2025gemini,
  title={Gemini 2.5: Pushing the frontier with advanced reasoning, multimodality, long context, and next generation agentic capabilities},
  author={Comanici, Gheorghe and Bieber, Eric and Schaekermann, Mike and Pasupat, Ice and Sachdeva, Noveen and Dhillon, Inderjit and Blistein, Marcel and Ram, Ori and Zhang, Dan and Rosen, Evan and others},
  journal={arXiv preprint arXiv:2507.06261},
  year={2025}
}

@misc{openai2025gptoss120bgptoss20bmodel,
      title={gpt-oss-120b and gpt-oss-20b Model Card}, 
      author={OpenAI},
      year={2025},
      eprint={2508.10925},
      archivePrefix={arXiv},
      primaryClass={cs.CL},
      url={https://arxiv.org/abs/2508.10925}, 
}

@misc{qwen2.5,
    title = {Qwen2.5: A Party of Foundation Models},
    url = {https://qwenlm.github.io/blog/qwen2.5/},
    author = {Qwen Team},
    month = {September},
    year = {2024}
}

@misc{seed2025seed-oss,
  author={ByteDance Seed Team},
  title={Seed-OSS Open-Source Models},
  year={2025},
  howpublished={\url{https://github.com/ByteDance-Seed/seed-oss}}
}

@misc{qwen3technicalreport,
      title={Qwen3 Technical Report}, 
      author={Qwen Team},
      year={2025},
      eprint={2505.09388},
      archivePrefix={arXiv},
      primaryClass={cs.CL},
      url={https://arxiv.org/abs/2505.09388}, 
}

@article{team2025kimi,
  title={Kimi k2: Open agentic intelligence},
  author={Team, Kimi and Bai, Yifan and Bao, Yiping and Charles, Y and Chen, Cheng and Chen, Guanduo and Chen, Haiting and Chen, Huarong and Chen, Jiahao and Chen, Ningxin and others},
  journal={arXiv preprint arXiv:2507.20534},
  year={2025}
}

@article{yue2026dr,
  title={Dr. Zero: Self-Evolving Search Agents without Training Data},
  author={Yue, Zhenrui and Upasani, Kartikeya and Yang, Xianjun and Ge, Suyu and Nie, Shaoliang and Mao, Yuning and Liu, Zhe and Wang, Dong},
  journal={arXiv preprint arXiv:2601.07055},
  year={2026}
}

@article{behrouz2025atlas,
  title={Atlas: Learning to optimally memorize the context at test time},
  author={Behrouz, Ali and Li, Zeman and Kacham, Praneeth and Daliri, Majid and Deng, Yuan and Zhong, Peilin and Razaviyayn, Meisam and Mirrokni, Vahab},
  journal={arXiv preprint arXiv:2505.23735},
  year={2025}
}

@article{behrouz2024titans,
  title={Titans: Learning to memorize at test time},
  author={Behrouz, Ali and Zhong, Peilin and Mirrokni, Vahab},
  journal={arXiv preprint arXiv:2501.00663},
  year={2024}
}

@article{behrouz2025nested,
  title={Nested learning: The illusion of deep learning architectures},
  author={Behrouz, Ali and Razaviyayn, Meisam and Zhong, Peilin and Mirrokni, Vahab},
  journal={arXiv preprint arXiv:2512.24695},
  year={2025}
}

@article{lu2026locas,
  title={Locas: Your Models are Principled Initializers of Locally-Supported Parametric Memories},
  author={Lu, Sidi and Liang, Zhenwen and Ma, Dongyang and Wang, Yan and Mi, Haitao and Yu, Dong},
  journal={arXiv preprint arXiv:2602.05085},
  year={2026}
}

@article{wang2024greater,
  title={With greater text comes greater necessity: Inference-time training helps long text generation},
  author={Wang, Yan and Ma, Dongyang and Cai, Deng},
  journal={arXiv preprint arXiv:2401.11504},
  year={2024}
}

@article{sun2024learning,
  title={Learning to (learn at test time): Rnns with expressive hidden states},
  author={Sun, Yu and Li, Xinhao and Dalal, Karan and Xu, Jiarui and Vikram, Arjun and Zhang, Genghan and Dubois, Yann and Chen, Xinlei and Wang, Xiaolong and Koyejo, Sanmi and others},
  journal={arXiv preprint arXiv:2407.04620},
  year={2024}
}

@article{moradi2025ttt,
  title={Continuous self-improvement of large language models by test-time training with verifier-driven sample selection},
  author={Moradi, Mohammad Mahdi and Amer, Hossam and Mudur, Sudhir and Zhang, Weiwei and Liu, Yang and Ahmed, Walid},
  journal={arXiv preprint arXiv:2505.19475},
  year={2025}
}

@article{hu2025ttl,
  title={Test-time learning for large language models},
  author={Hu, Jinwu and Zhang, Zhitian and Chen, Guohao and Wen, Xutao and Shuai, Chao and Luo, Wei and Xiao, Bin and Li, Yuanqing and Tan, Mingkui},
  journal={arXiv preprint arXiv:2505.20633},
  year={2025}
}

@inproceedings{sun2020test,
  title={Test-time training with self-supervision for generalization under distribution shifts},
  author={Sun, Yu and Wang, Xiaolong and Liu, Zhuang and Miller, John and Efros, Alexei and Hardt, Moritz},
  booktitle={International conference on machine learning},
  pages={9229--9248},
  year={2020},
  organization={PMLR}
}

@article{liu2026test,
  title={Test-Time Training with KV Binding Is Secretly Linear Attention},
  author={Liu, Junchen and Elflein, Sven and Litany, Or and Gojcic, Zan and Li, Ruilong},
  journal={arXiv preprint arXiv:2602.21204},
  year={2026}
}

@inproceedings{kwon2023efficient,
  title={Efficient memory management for large language model serving with pagedattention},
  author={Kwon, Woosuk and Li, Zhuohan and Zhuang, Siyuan and Sheng, Ying and Zheng, Lianmin and Yu, Cody Hao and Gonzalez, Joseph and Zhang, Hao and Stoica, Ion},
  booktitle={Proceedings of the 29th symposium on operating systems principles},
  pages={611--626},
  year={2023}
}

@inproceedings{aminabadi2022deepspeed,
  title={Deepspeed-inference: enabling efficient inference of transformer models at unprecedented scale},
  author={Aminabadi, Reza Yazdani and Rajbhandari, Samyam and Awan, Ammar Ahmad and Li, Cheng and Li, Du and Zheng, Elton and Ruwase, Olatunji and Smith, Shaden and Zhang, Minjia and Rasley, Jeff and others},
  booktitle={SC22: International Conference for High Performance Computing, Networking, Storage and Analysis},
  pages={1--15},
  year={2022},
  organization={IEEE}
}

@article{wang2025explore,
  title={Explore to Evolve: Scaling Evolved Aggregation Logic via Proactive Online Exploration for Deep Research Agents},
  author={Wang, Rui and Zhang, Ce and Ma, Jun-Yu and Zhang, Jianshu and Wang, Hongru and Chen, Yi and Xue, Boyang and Fang, Tianqing and Zhang, Zhisong and Zhang, Hongming and others},
  journal={arXiv preprint arXiv:2510.14438},
  year={2025}
}

@article{wan2026inference,
  title={Inference-Time Scaling of Verification: Self-Evolving Deep Research Agents via Test-Time Rubric-Guided Verification},
  author={Wan, Yuxuan and Fang, Tianqing and Li, Zaitang and Huo, Yintong and Wang, Wenxuan and Mi, Haitao and Yu, Dong and Lyu, Michael R},
  journal={arXiv preprint arXiv:2601.15808},
  year={2026}
}

\appendix

\clearpage

\section{Details of Input Processing}\label{appendix:input_processing}

To reduce noise in large-scale web data and focus on domain-relevant information, we perform \textbf{importance scoring} and \textbf{clustering} over webpages.

\mybold{Importance Scoring.} We model a website as a directed graph, where each node represents a webpage and a directed edge \(A \rightarrow B\) indicates that page \(A\) links to page \(B\). Let \(d_{\text{in}}(v)\) and \(d_{\text{out}}(v)\) denote the in-degree and out-degree of node \(v\), respectively. We define the importance of a node as:
\[
\text{Importance}(v) = 0.7 \cdot d_{\text{in}}(v) + 0.3 \cdot d_{\text{out}}(v).
\]

\mybold{Clustering.} To impose structure on complex websites, we group webpages into clusters based on shared URL prefixes. Starting from the first path segment, URLs are recursively partitioned until each group satisfies a size constraint.  This strategy organizes webpages into coherent categories, making the overall structure more interpretable and easier to navigate for downstream processing.

\section{Example Chowcase}\label{appendix:example}


\begin{case_data}{An Example of the Processed Input}{exmp:agent}

{\ttfamily\small\obeylines
Total: 5 clusters, 221 URLs | per-cluster sizes: [40, 10, 130, 21, 20]
============================================================

[Prefix] https://sigchi.org (40 URLs)
https://sigchi.org/conferences/upcoming [score:227]
https://sigchi.org/people/all-committees [score:226]
https://sigchi.org/news/announcements [score:223]
https://sigchi.org [score:222]
https://sigchi.org/people/sigchi-awards [score:222]
... (35 URLs omitted)

============================================================

[Prefix] https://sigchi.org/about/policies (10 URLs)
https://sigchi.org/about/policies [score:20]
https://sigchi.org/about/policies/conference-policies [score:17]
https://sigchi.org/about/policies/conference-policies/data [score:12]
https://sigchi.org/about/policies/conference-policies/submission-and-review [score:11]
https://sigchi.org/about/policies/conference-policies/conference-guidelines [score:11]
... (5 URLs omitted)

============================================================

[Prefix] https://sigchi.org/events (130 URLs)
https://sigchi.org/events [score:140]
https://sigchi.org/events/open-session-on-acm-open [score:11]
https://sigchi.org/events/ahs26 [score:11]
https://sigchi.org/events/september-coffee-hour [score:11]
https://sigchi.org/events/sigchi-town-hall [score:11]
... (125 URLs omitted)

============================================================

[Prefix] https://sigchi.org/people/committees (21 URLs)
https://sigchi.org/people/committees/asia-committee [score:11]
https://sigchi.org/people/committees/asia-committee/bylaws [score:11]
https://sigchi.org/people/committees/research-ethics-committee [score:11]
https://sigchi.org/people/committees/awards-committee [score:11]
https://sigchi.org/people/committees/development-fund-committee [score:11]
... (16 URLs omitted)

============================================================

[Prefix] https://sigchi.org/resources (20 URLs)
https://sigchi.org/resources/guides-for-organizers [score:221]
https://sigchi.org/resources/gary-marsden-travel-awards [score:220]
https://sigchi.org/resources/sigchi-development-fund [score:220]
https://sigchi.org/resources/guides-for-authors [score:220]
https://sigchi.org/resources/sigchi-development-fund/recipients [score:15]
... (15 URLs omitted)
}

\end{case_data}

\begin{case_data}{An Example of World Knowledge}{exmp:agent}
\textbf{\large Overview}
\begin{itemize}
    \item \textbf{Website:} https://2024.aclweb.org
    \item \textbf{Total Categories:} 2
    \item \textbf{Total Pages Analyzed:} 31
\end{itemize}
The ACL 2024 website serves as the official information hub for the 62nd Annual Meeting of the Association for Computational Linguistics. It provides comprehensive details for participants, including registration, program schedules, venue information, and various calls for participation.

\noindent\rule{\textwidth}{0.4pt}

\vspace{1em}
\textbf{\large Category: ACL 2024 Main Website}
\begin{itemize}
    \item \textbf{URL Prefix:} https://2024.aclweb.org
    \item \textbf{Category Summary:} This category covers the main pages of the ACL 2024 conference website. It includes essential information for attendees, such as registration details, program schedules, venue information, and calls for participation. These pages serve as the primary resource for anyone interested in or attending the conference.
\end{itemize}

\textbf{Scraped Pages:}
\begin{itemize}
    \item \textbf{Main Conference - ACL 2024}: The 2024 Annual Meeting of the Association for Computational Linguistics (ACL) will be held in Bangkok, Thailand from August 11-16, 2024. The conference's special theme is "Open science, open data, and open models for reproducible NLP research," and it covers a wide range of topics in Natural Language Processing. Paper submissions for the main conference were due on February 15, 2024, via the ACL Rolling Review (ARR) system. URL: https://2024.aclweb.org/calls/main\_conference\_papers/
    \item \textbf{Participants - ACL 2024}: This page provides logistical information for conference participants, focusing on the venue and accommodations. The conference is held at the Centara Grand and Bangkok Convention Centre in Thailand, and the page lists several hotel options with rates. It is noted that many of the listed hotels are already fully booked, and the page does not contain information on dates, deadlines, or topics. URL: https://2024.aclweb.org/participants/
    \item \textbf{Frequently Asked Questions - ACL 2024}: This page provides key logistical information for ACL 2024 attendees, detailing presentation formats such as 10-minute talks for long papers and poster-only sessions for Findings papers. It also serves as a central resource with important deadlines and provides essential links for the conference schedule, visa invitation letters, on-site poster printing, and paper submissions. The page clarifies that while Findings papers don't require a presentation, author registration is still mandatory. URL: https://2024.aclweb.org/faq/
    \item \textbf{Sponsors - ACL 2024}: This page details the corporate sponsors for the ACL 2024 conference, noting that the deadline for sponsorship is June 30. The sponsors are organized into five distinct tiers, from Diamond to Bronze, based on their level of contribution. Major technology companies like Apple, Google Deepmind, Meta, and Amazon are listed as top-tier Diamond and Platinum sponsors. URL: https://2024.aclweb.org/sponsors
    \item \textbf{Organizing Committee - ACL 2024}: The ACL 2024 conference will take place in Bangkok, Thailand, from August 11-16, 2024, with a special theme on "Open science, open data, and open models for reproducible NLP research." The provided text details important dates for the event, including a paper submission deadline of February 15, 2024, and a notification of acceptance on May 15, 2024. It also lists a wide range of submission topics, such as Machine Translation, Ethics, Dialogue Systems, and NLP Applications. URL: https://2024.aclweb.org/organization/
    \item \textbf{Registration - ACL 2024}: This page outlines the registration options for the ACL 2024 conference, including in-person, virtual, and tutorial/workshop-only passes. Key details include that authors must register their papers by July 12, 2024, and that discounted virtual registration is available for those with financial need. In-person attendance includes full access to sessions and meals, while virtual access is provided through the Underline platform. URL: https://2024.aclweb.org/registration
    \item \textbf{Visa - ACL 2024}: This page provides essential visa information for attendees of the ACL 2024 conference traveling to Thailand, directing them to the official Thai e-visa website. It specifies that applicants should select 'Participants of MICE Industry' as their purpose of visit and lists the various countries whose citizens are eligible for visa exemption or visa on arrival. The page also includes links to official resources and special instructions for certain nationalities. URL: https://2024.aclweb.org/participants/visa/
    \item \textbf{Printing Service - ACL 2024}: The ACL 2024 conference offers a printing service for participant posters, recommending an A0 size. Attendees can secure an early rate of THB 1,400 (\textasciitilde USD 40) by ordering before the August 5 deadline. The price doubles for the normal rate (until August 9) and increases substantially for onsite orders, so participants are strongly encouraged to print in advance to save money. URL: https://2024.aclweb.org/participants/printing/
    \item \textbf{D\&I Socials and BoF Sessions - ACL 2024}: This page is a call for individuals or groups interested in organizing Birds of a Feather (BoF) sessions or affinity group meetings for the ACL 2024 conference. To express interest and provide feedback, parties are instructed to fill out a form by the deadline of July 31, 2024. For any inquiries, the ACL 2024 Diversity and Inclusion team can be contacted via email. URL: https://2024.aclweb.org/calls/bof/
    \item \textbf{Student Research Workshop - ACL 2024}: The ACL 2024 Student Research Workshop is a forum for students to present their work, accepting submissions for both research papers and thesis proposals. The workshop offers an optional pre-submission mentorship program and financial grants to support student attendance, with the final paper submission deadline set for June 1, 2024. URL: https://2024.aclweb.org/calls/srw/
    \item \textbf{Venue - ACL 2024}: The ACL 2024 conference will be held in Bangkok, Thailand, and this page provides essential information for attendees. It offers practical travel advice covering the city's tropical climate, the local currency (Thai Baht), and general safety precautions. The page also includes important resources like emergency contact numbers and suggestions for popular tourist destinations to help visitors prepare for their trip. URL: https://2024.aclweb.org/participants/venue/
    \item \textbf{Student Volunteer Program - ACL 2024}: The ACL 2024 conference is accepting applications for its Student Volunteer Program until the deadline of July 3, 2024. Accepted full-time students receive benefits like free conference registration and an ACL membership, but are responsible for their own travel and accommodation costs. Volunteers are selected based on enthusiasm, financial need, and whether they are presenting a paper, and they must fulfill all assigned duties to avoid being charged the full registration fee. URL: https://2024.aclweb.org/calls/volunteers/
    \item \textbf{Diversity and Inclusion Subsidies - ACL 2024}: ACL 2024 is offering various subsidies, including funds for registration, caregiving, and travel, to support researchers from developing countries, marginalized communities, students, and those with financial hurdles. Applications are due by July 3rd, 2024, with preference given to first-time attendees, presenters from underrepresented groups, and those without other financial support. URL: https://2024.aclweb.org/calls/diversity\_and\_inclusion\_subsidies/
    \item \textbf{System Demonstrations - ACL 2024}: This page details the call for system demonstration submissions for the ACL 2024 conference, covering a range of topics including NLP systems, software tools, and data visualization. Submissions, due by March 18, 2024, are limited to a 6-page paper and must be accompanied by a short video demonstrating the system. The review process is single-blind, and a "Best Demo Award" will be presented to an outstanding submission. URL: https://2024.aclweb.org/calls/system\_demonstrations/
    \item \textbf{Ethical Policies - ACL 2024}: This page outlines the ethical guidelines for the ACL 2024 conference, emphasizing its adherence to the ACL Anti-Harassment Policy. It underscores the collective responsibility of the community to create an inclusive and positive atmosphere. Participants are provided with clear instructions to confidentially report any harassment or hostile behavior to the ACL Professional Conduct Committee for consultation and action. URL: https://2024.aclweb.org/participants/ethical\_policies/
    \item \textbf{Travel - ACL 2024}: This page provides essential travel information for individuals attending the ACL 2024 conference in Bangkok. It outlines details on the city's two main airports (BKK and DMK), lists local transportation options like the recommended BTS SkyTrain, and confirms that complimentary parking is available for attendees. URL: https://2024.aclweb.org/participants/travel/
\end{itemize}

\begin{quote}
This category may contain additional pages beyond those listed. For further exploration, visit: https://2024.aclweb.org
\end{quote}

\vspace{1em}
\textbf{\large Category: ACL 2024 Program Information}
\begin{itemize}
    \item \textbf{URL Prefix:} https://2024.aclweb.org/program
    \item \textbf{Category Summary:} This section details the conference program for ACL 2024. It includes links to the overall schedule, specific tracks, workshops, tutorials, and other related events. This is the central hub for attendees to plan their schedule and find information about the various sessions.
\end{itemize}

\textbf{Scraped Pages:}
\begin{itemize}
    \item \textbf{Conference Overview - ACL 2024}: The ACL 2024 conference is scheduled from August 10th to August 16th, featuring a comprehensive program that includes tutorials, workshops, and various presentation formats. The event will host keynote speeches by prominent researchers Sunita Sarawagi, Subbarao Kambhampati, and Barbara Plank, and provides access to the titles of all papers from the main conference, findings, and demo tracks. URL: https://2024.aclweb.org/program/
    \item \textbf{Accepted Main Conference Papers - ACL 2024}: This page provides a comprehensive list of all papers accepted to the ACL 2024 main conference, separated into 'Long Papers' and 'Short Papers' categories. Each listed entry includes the full paper title and the names of the respective authors. The page functions solely as a list and does not provide additional details such as presentation schedules, topics, or other instructions. URL: https://2024.aclweb.org/program/accepted\_main\_conference\_papers/
    \item \textbf{Accepted SRW Papers - ACL 2024}: This page provides a complete list of accepted papers for the Student Research Workshop (SRW) at the ACL 2024 conference. The accepted works are divided into two categories, "Oral Papers" and "Poster Papers," with each entry listing the paper's title and the names of the authors. The page exclusively serves as a list and does not contain presentation schedules or group the papers by topic. URL: https://2024.aclweb.org/program/accepted\_srw\_papers/
    \item \textbf{Keynotes - ACL 2024}: The page introduces the three keynote speakers for the ACL 2024 conference: Sunita Sarawagi, Subbarao Kambhampati, and Barbara Plank. Their respective talks will cover pressing topics in AI, such as the tradeoffs of in-context-learning for model adaptation, whether LLMs can truly reason and plan, and the need to embrace variation in NLP to avoid narrowing the field's horizons. URL: https://2024.aclweb.org/program/keynotes/
    \item \textbf{Tutorials - ACL 2024}: The ACL 2024 conference will host six tutorials divided into morning (09:00 - 12:30) and afternoon (14:00 - 17:30) sessions. Topics focus heavily on large language models, covering their vulnerabilities, watermarking, and expressivity, alongside other key areas like computational linguistics for brain decoding and effective science communication. Specific tutorials also address automatic text generation and revision. URL: https://2024.aclweb.org/program/tutorials/
    \item \textbf{Panel - ACL 2024}: The ACL 2024 conference will host a panel titled ``Challenges and Opportunities with SEA LLMs'' to address the development of Large Language Models for low-resource Southeast Asian languages. Chaired by Lun-Wei Ku, the session features experts Dr. Sarana Nutanong (VISTEC), Dr. Ayu Purwarianti (ITB), and Dr. William Tjhi (AI Singapore). Key discussion points will include data requirements, application scenarios, and strategies for creating and maintaining LLMs for the region. URL: https://2024.aclweb.org/program/panel/
    \item \textbf{Best Paper Awards - ACL 2024}: This page lists the award-winning papers from the ACL 2024 conference, divided into several categories including Best Paper, Best Social Impact, Best Resource, and Best Theme. The three Best Paper awards were given to 'Mission: Impossible Language Models', 'Semisupervised Neural Proto-Language Reconstruction', and 'Why are Sensitive Functions Hard for Transformers?'. The page also highlights winners in other categories, such as 'DIALECTBENCH' for social impact and 'Dolma' for best resource, along with lists of Outstanding Papers and SAC Awards. URL: https://2024.aclweb.org/program/best\_paper\_awards/
    \item \textbf{Accepted CL Papers - ACL 2024}: This page provides the official list of accepted papers for the Computational Linguistics (CL) track of the ACL 2024 conference. Each entry includes the full paper title and the names of its authors, such as 'Stance Detection with Explanations' by Rudra Ranajee Saha et al. The page functions solely as a list and does not contain conference schedules, paper topics, or presentation instructions. URL: https://2024.aclweb.org/program/accepted\_cl\_papers/
    \item \textbf{Accepted Findings Papers - ACL 2024}: This page lists the accepted "Findings" research papers for the ACL 2024 conference. The content is organized into two distinct sections, 'Long Papers' and 'Short Papers', with each entry providing the paper's title and its authors. The page serves as a directory for these specific papers but does not include information on schedules or topics. URL: https://2024.aclweb.org/program/accepted\_findings\_papers/
    \item \textbf{Workshops - ACL 2024}: This page lists the workshops for the ACL 2024 conference, held on August 15 and 16, 2024. The workshops cover a wide variety of specific topics in Natural Language Processing, such as argument mining, machine translation for low-resource languages, and social media mining for health research. Each listing provides key details including the workshop's title, date, location, and organizers. URL: https://2024.aclweb.org/program/workshops/
\end{itemize}

\begin{quote}
This category may contain additional pages beyond those listed. For further exploration, visit: https://2024.aclweb.org/program
\end{quote}

\end{case_data}

\section{Prompt Showcase}\label{appendix:prompt}

\begin{exmp}{World Knowledge Generation Prompt (For teacher agent)}{exmp:agent}

\textbf{Role}

You are a Web Intelligence Agent specializing in website analysis and knowledge organization.
You will receive a \textbf{pre-clustered URL file} for a website, where URLs are already grouped by path prefix. Your task is to scrape these URLs \textbf{category by category} and produce a structured \textbf{World Knowledge} that stays within a target token range.

\textbf{Constraints}
\begin{itemize}
    \item \textbf{Maximum} World Knowledge length: \textbf{\{token\_limit\}} tokens
    \item \textbf{Minimum} World Knowledge length: \textbf{\{min\_token\}} tokens
    \item You must actively manage content length throughout the process — compress when too long, expand when too short.
    \item \textbf{No external links:} Do NOT include any \textbf{external links} that lead to a different domain in the Guidebook. Only document pages belonging to the website's own domain.
    \item \textbf{Summarize, don't copy:} Do NOT copy raw page content verbatim into the Guidebook. Always \textbf{summarize and condense} the key information in your own words. The Guidebook should be a concise guide, not a dump of webpage text.
    \item \textbf{Every scraped page must have a URL:} Each entry under "Scraped Pages" \textbf{MUST} include the full URL in parentheses. An entry without a URL is \textbf{INVALID}. Format: \texttt{- \textbf{[Page Title]} : [summary]}. Never write a summary without its corresponding URL.
    \item \textbf{Follow your token plan:} The length and detail of each category's content should be guided by the \textbf{planned token allocation} from Phase 0. Spend more tokens on categories with more URLs, fewer on small ones. Focus on specific, useful information (names, dates, numbers, features). Do not pad with generic or repetitive descriptions.
\end{itemize}

\textbf{URL Priority Rules} 

ach URL in the cluster file has a structure-based score in the format \texttt{[score:S]}. You must evaluate the importance of each webpage by combining this structural score with its semantic relevance and content value. Use this combined assessment to determine the page's overall priority, whether it should be selected, and how much detail (token budget) to dedicate to it in your subsequent writing.

\textbf{Input}\\
A clustered URL file is located at: \texttt{\{queue\_file\_path\}}

The file format:
\begin{verbatim}
Total: <N> clusters, <M> URLs |  per-cluster sizes: [<c1>, <c2>, ...]
============================================================
[Prefix] <prefix_url>  (<total> URLs)
<url_1>  [score:<S>]
<url_2>  [score:<S>]
...
============================================================
[Prefix] <prefix_url>  (<total> URLs)
<url_3>  [score:<S>]
<url_4>  [score:<S>]
...

\end{verbatim}

The first line contains global statistics: total number of clusters, total number of URLs and the URL count for each cluster after \texttt{per-cluster sizes:}. Categories are separated by lines of repeated \texttt{=} characters. Each category starts with a \texttt{[Prefix]} header showing \texttt{(shown/total URLs)}. Each URL is annotated with \texttt{[score:S]}. \texttt{score} is a composite importance score. 

\textbf{Tools}\\
To assist you with this task, I have provided a complete Python code block below containing the necessary tool functions. Please make sure to copy and run this entire block as your first step. After that, you can conveniently use the functions by calling their names. It is highly recommended to use them as-is without rewriting or re-implementing them, unless you run into a runtime error that requires a modification.

\begin{verbatim}
{tool_functions_code}
\end{verbatim}

\textbf{Workflow}

Please follow the phases in their natural order: Phase 0 $\rightarrow$ Phase 1 $\rightarrow$ Phase 2. It is highly recommended to complete Phase 0 first, as Phase 1 relies on the plan file created during this initial step. Jumping straight to Phase 1 will cause \texttt{read\_plan()} to fail and require unnecessary recovery steps.

\subsubsection*{Phase 0: Initialization \& Planning}

Before processing any category, please create a token-allocation plan to guide your work.

\textbf{Step 1: Parse Cluster Statistics}
Call \texttt{parse\_cluster\_stats()}. It reads the first line of the queue file and extracts: Total number of clusters, total number of URLs, and item Per-cluster URL counts.

\textbf{Step 2: Create Token Allocation Plan}
\begin{itemize}
    \item \textbf{Allocate tokens proportionally by effective URL count} — more URLs $\rightarrow$ more tokens. Try to avoid splitting them equally. Rough guide: $\sim$50–80 tokens per page entry + $\sim$100–150 for category header/summary.
    \item \textbf{Please ensure the total planned tokens fall within [\{min\_token\}, \{token\_limit\}].} Adjust your allocations as needed to stay in this range.
    \item Call \texttt{write\_plan(plan\_text)} to save to \texttt{\{plan\_file\_path\}}. 
\end{itemize}

\textbf{Step 3:} Proceed to Phase 1.

---

\subsubsection*{Phase 1: Category-by-Category Processing (Loop)}

Process the clustered URL file one category at a time:

\textbf{Step 1: Load Next Category}
\begin{itemize}
    \item Call \texttt{get\_next\_category()} — the one already defined above. This function returns the current unprocessed category block. It is safe to call multiple times — it always returns the same category until you explicitly call \texttt{mark\_category\_done()}.
    \item If it returns \texttt{None}, all categories have been processed — proceed to \textbf{Phase 2}.
\end{itemize}

\textbf{Step 1.5: Read Token Budget (MANDATORY — do NOT skip)}
\begin{itemize}
    \item \textbf{This step is NOT optional. You MUST execute it for every category before writing anything.}
    \item Call \texttt{read\_plan()} to open and read the plan file (\texttt{\{plan\_file\_path\}}).
    \item If \texttt{read\_plan()} returns an empty string, \textbf{STOP and go back to Phase 0 Step 2 to create the plan first.} Never proceed without a plan.
    \item Find the current category's prefix URL in the plan and extract its \textbf{planned token allocation} (e.g., \texttt{budget = 1200}).
    \item \textbf{Print it explicitly:} \texttt{print(f"Token budget for this category: \{budget\}")} — this forces you to be aware of the target.
    \item You will use this number in Step 3 to control the length of your output. For example: if the budget is 500 tokens, write a brief summary with short page entries; if 2000 tokens, write detailed summaries with rich page entries.
\end{itemize}

\textbf{Step 2: Select \& Scrape Member Pages}
\begin{itemize}
    \item Please selectively choose which URLs to scrape based on their relative importance and your available token limits. 
    \item To help prioritize the most valuable pages, you can refer to the \texttt{[score:S]} metric provided for each URL, selecting those with higher scores first.
    \item For each selected URL, fetch the webpage and extract the key information.
    \item To make the best use of your tokens, please skip pages that do not provide meaningful content (such as login walls, error pages, empty pages, or cookie/privacy notices).
\end{itemize}

\textbf{Step 3: Write Category Section (target the token budget from Step 1.5)}
\begin{itemize}
    \item \textbf{Keep your token budget in mind:} Aim to keep the entire section (header, summary, and scraped entries) close to the budget you extracted in Step 1.5. A variance of around $\pm$20\% is perfectly fine.
    \item Use \texttt{append\_to\_guidebook(text)} to add this category's section to the guidebook.
    \item \textbf{Category Summary:} Briefly describe the main topics and types of pages found here, adjusting the level of detail to fit your budget.
    \item \textbf{Formatting:} Please follow the template below closely. It is important to keep the exact header structure (\texttt{\#\# Category: [Name]}).
\begin{verbatim}
## Category: [Descriptive Name Based on Content]
- **URL Prefix:** [the prefix URL for this group]
- **Category Summary:** [Describe the main topics.]

**Scraped Pages:**
- **[Page Title]** ([full URL]): [Specific details like names, dates, 
  or features. Adjust length to fit budget.]
- **[Page Title]** ([full URL]): [summary]
- ...

> This category may contain additional pages beyond those listed. 
  For further exploration, visit: [prefix URL]
\end{verbatim}
    \item \textbf{Include URLs:} Please ensure every entry under "Scraped Pages" includes its full URL in parentheses.
\end{itemize}

\textbf{Step 4: Mark Done \& Continue}
\begin{itemize}
    \item After successfully appending this category to the guidebook, call \texttt{mark\_category\_done()} to advance the progress index. \textbf{Only call this after \texttt{append\_to\_guidebook()} succeeds} — this ensures no category is skipped even if an earlier step fails or retries.
    \item Return to \textbf{Step 1} to process the next category.
\end{itemize}

---

\subsubsection*{Phase 2: Refinement \& Finalization}

After all categories have been processed:

\textbf{Step 1: Token-Based Compression or Expansion}
\begin{itemize}
    \item Call \texttt{count\_guidebook\_tokens()}.
    \item \textbf{If tokens $>$ \{token\_limit\}:} Compress the guidebook:
    \begin{itemize}
        \item Call \texttt{read\_guidebook()} to review all content.
        \item Identify verbose or repetitive sections and rewrite them with \texttt{rewrite\_category\_section()}.
        \item Repeat until within limit.
    \end{itemize}
    \item \textbf{If tokens $<$ \{min\_token\}:} Expand the guidebook. Follow these steps \textbf{exactly} to avoid content loss:
\begin{enumerate}
    \item \textbf{Identify areas for expansion:} Review your plan and the current guidebook. Feel free to choose any category that seems underdeveloped or has interesting URLs you haven't explored yet.
    \item \textbf{Review existing content:} Use \texttt{read\_guidebook()} to see what has already been written for your chosen category.
    \item \textbf{Explore and gather new data:} Scrape additional URLs within that category to discover fresh details. Please rely on the actual webpage content to inspire your expansion and ensure accuracy.
    \item \textbf{Integrate and enrich:} Seamlessly weave your new discoveries into the existing text. You can expand summaries, add new page entries, or provide deeper insights to make the section richer and more comprehensive.
    \item \textbf{Update the guidebook:} Use the \texttt{rewrite\_category\_section(category\_name, new\_section\_text)} function to replace the old section with your newly expanded version.
    \item \textbf{Check progress:} Use \texttt{count\_guidebook\_tokens()} to see how close you are to your goal. Continue this exploration process until your guidebook reaches at least \texttt{\{min\_token\}} tokens.
\end{enumerate}

\end{itemize}

\textbf{Step 2: Add Overview Header \& Save}
\begin{itemize}
    \item Call \texttt{read\_guidebook()} to get the full current content.
    \item Prepend an Overview section at the top:
\begin{verbatim}
# [Website Domain] Guidebook

## Overview
- **Website:** [base URL]
- **Total Categories:** [number]
- **Total Pages Analyzed:** [number]
[2-3 sentence high-level overview of the website's purpose and content.]

---
\end{verbatim}
    \item Call \texttt{save\_final\_guidebook(full\_content)} with the complete content (Overview + all category sections).
\end{itemize}

\end{exmp}

\begin{exmp}{World Knowledge Generation Prompt (For trained agents)}{exmp:web_agent}
You are a Web Intelligence Agent that analyzes websites and organizes their content into a structured knowledge document called a \textbf{Guidebook} — a concise, categorized summary of a website's pages and their key information.

\vspace{0.5em}
\textbf{Input} \\
Your input is a \textbf{clustered URL file} at \texttt{\{queue\_file\_path\}}. This file contains URLs from a single website, pre-grouped into categories by their URL path prefix (e.g., all \texttt{/blog/...} URLs form one category, all \texttt{/docs/...} URLs form another). Each URL is annotated with link metrics in the format \texttt{[in:X out:Y score:S]}, where \texttt{in} is how many other pages link to it (inbound links), \texttt{out} is how many links it contains (outbound links), and \texttt{score} is a composite importance score derived from both. Categories are separated by \texttt{===...===} lines, and each starts with a \texttt{[Prefix]} header.

\vspace{0.5em}
\textbf{Tools} \\
You have access to \texttt{web\_agent(task=...)}, a function that fetches and reads real web pages — use it to scrape each URL's content. In addition, the code block below provides helper functions for managing the Guidebook (appending content, tracking progress, counting tokens, etc.). Copy and execute this entire block in your first code cell:

\vspace{0.5em}
\texttt{\{tool\_functions\_code\}}

\vspace{0.5em}
\textbf{Tool Usage}
\begin{enumerate}
    \item Call \texttt{parse\_cluster\_stats()} to read the file header and get the total number of URLs and categories. Based on the site size, decide your processing mode: for small sites ($\le$ 250 URLs), use \textbf{FULL mode} where every URL is included; for larger sites, use \textbf{SELECTIVE mode} where you pick the most important URLs per category (ranked by \texttt{score}, up to 20 per category if $\le$ 8 categories, or 10 if more).
    \item Create a token allocation plan — distribute the target Guidebook length (\texttt{\{min\_token\}}–\texttt{\{token\_limit\}} tokens) across categories proportionally by each category's \textbf{effective URL count} (i.e., the number of URLs you will actually scrape, after applying the per-category cap from step 1 — not the raw total), then save it with \texttt{write\_plan()}.
    \item Process categories one by one: call \texttt{get\_next\_category()} to load a category, scrape its selected URLs with \texttt{web\_agent()}, write the category section with \texttt{append\_to\_guidebook()}, then call \texttt{mark\_category\_done()} to advance. Repeat until all categories are done.
    \item After all categories are processed, check the total length with \texttt{count\_guidebook\_tokens()}. If it exceeds \texttt{\{token\_limit\}}, compress verbose sections with \texttt{rewrite\_category\_section()}. If it falls below \texttt{\{min\_token\}}, expand by scraping additional URLs. Finally, prepend an Overview header and call \texttt{save\_final\_guidebook()}.
\end{enumerate}

\vspace{0.5em}
\textbf{Output format per category:}
\begin{quote}
\ttfamily
\#\# Category: [Name] \\
- **URL Prefix:** [prefix URL] \\
- **Category Summary:** [what this category covers] \\
\\
**Scraped Pages:** \\
- **[Page Title]** : [summary of key info] \\
- ... \\
\\
> This category may contain additional pages. Visit: [prefix URL]
\end{quote}

\vspace{0.5em}
\textbf{Rules:}
\begin{itemize}
    \item Scraping: You MUST call \texttt{web\_agent(task="..."{})} to fetch real content for every selected URL. Summarize the key information. Crucially, every page summary must come from a real \texttt{web\_agent()} call. Never fabricate or guess content. Do NOT use placeholders to stand in for URL summaries, and NEVER rely on your internal knowledge to hallucinate or invent page content.
    \item Every scraped page entry must include its full URL.
    \item Only include pages from the website's own domain — no external links.
    \item Summarize in your own words; do not copy page content verbatim.
    \item Process ALL categories before finalizing.
\end{itemize}
\end{exmp}

\begin{exmp}{Evaluation Prompt for LLM Judge (WebWalker)}{exmp:agent}
Your task is to determine whether the answer is consistent with the ground truth for the given question.

\textbf{Evaluation rules:}
\begin{enumerate}
    \item Output 1 if the answer correctly answers the question and has the same meaning as the ground truth.
    \item The answer does NOT need to exactly match the ground truth.
    \item Differences in wording, format, order, or level of detail are acceptable as long as the meaning is equivalent.
    \item Concise answers should NOT be judged as incorrect simply because they are shorter than the ground truth.
    \item Different formats that express the same information (e.g., numbers only, different date formats, paraphrases) should be considered correct.
    \item Output 0 only if the answer is incorrect, contradicts the ground truth, or fails to answer the question.
\end{enumerate}

\textbf{Examples:}

\textbf{Example 1} \\
\textbf{Question:} What are the 2024 suggested retail prices of the Yamaha PAC612 electric guitar and the Sonogenic SHS-300 shoulder keyboard? \\
\textbf{Ground truth:} PAC612 electric guitar suggested retail price: 8,400 RMB. SHS-300 shoulder keyboard suggested retail price: 1,299 RMB (white) and 1,399 RMB (blue). \\
\textbf{Answer:} 8400,1299,1399 \\
\textbf{Judgment:} 1

\vspace{0.5em}
\textbf{Example 2} \\
\textbf{Question:} What is Jack's birthday? \\
\textbf{Ground truth:} December 10 \\
\textbf{Answer:} 12-10 \\
\textbf{Judgment:} 1

\vspace{0.5em}
\textbf{Example 3 (Prompt Rules Evaluation)} \\
\textbf{Question:} What are the conditions for outputting 1 according to the evaluation rules? \\
\textbf{Ground truth:} Output 1 if the answer correctly answers the question and has the same meaning as the ground truth. Differences in wording, format, order, or level of detail are acceptable. Concise answers or different formats expressing the same information are also correct. \\
\textbf{Answer:} Give it a 1 if the core meaning matches, even if the answer is shorter, formatted differently, or paraphrased. \\
\textbf{Judgment:} 1

\vspace{1em}
Now evaluate the following case.

\textbf{Question:} \texttt{\{question\}} \\
\textbf{Answer:} \texttt{\{predict\}} \\
\textbf{Ground truth:} \texttt{\{gt\}}

Output only one number: \\
1 if the answer is correct or semantically equivalent to the ground truth, otherwise 0. \\
Do not output anything other than the number.
\end{exmp}

\begin{exmp}{Evaluation Prompt for LLM Judge (WebVoyager)}{exmp:with_trees}

Your task is to determine whether the web task has been successfully accomplished, based on the task instruction, the result response, and the accessibility trees of the webpages.

You are given three components:
\begin{enumerate}
    \item Web Task Instruction: A natural language instruction describing the task to be completed (e.g., search, verify, compare, summarize).
    \item Result Response: The final textual response generated after performing the task.
    \item Accessibility Trees: Structured representations of the webpages at each step, serving as evidence of the actions taken.
\end{enumerate}

\textbf{Evaluation rules:}
\begin{enumerate}
    \item You do NOT need to interact with websites or perform any real actions.
    \item You must base your judgment only on the provided instruction, response, and accessibility trees. Do NOT assume missing information.
    \item Your primary goal is to evaluate whether the actions reflected in the trees and the final response correctly follow the instruction.
    \item If the task contains multiple requirements (e.g., find information and summarize it), all must be completed. Missing any part leads to NOT SUCCESS.
    \item The accessibility trees serve as ground truth evidence of what actually happened during execution.
    \item If the Result Response contradicts the trees, trust the trees.
    \item If the Result Response contains information not present in the trees, trust the response.
\end{enumerate}

\textbf{Instructions:}
You should briefly explain your reasoning before giving the final verdict.

\vspace{0.5em}
Now evaluate the following case.

\textbf{TASK:} \texttt{\{task\}} \\
\textbf{Result Response:} \texttt{\{answer\}} \\
\textbf{Accessibility Trees:} \texttt{\{trees\}}

\vspace{0.5em}
Output your final verdict as one of the following:

\texttt{SUCCESS} \\
\texttt{NOT SUCCESS}

Do not output anything other than the verdict.

\end{exmp}

\begin{exmp}{Query Generation Prompt}{exmp:query_generation}
You are an advanced web information gathering expert and navigation planner.
I will provide you with a [Main Page URL], a [World Knowledge] containing summaries of sub-pages, and a [Target Question] that needs to be resolved.
\textbf{Your task is to evaluate this information, determine which specific web pages should be explored next, and ultimately obtain the final answer to the question by visiting these pages.}

\vspace{0.5em}
\textbf{[Input Data]}
\begin{enumerate}
    \item Main Page URL: \texttt{\{URL\}}
    \item Target Question: \texttt{\{Question\}}
    \item Sub-page Summaries: \texttt{\{World Knowledge\}}
\end{enumerate}

\vspace{0.5em}
\textbf{[Your Decision Logic]} \\
Please carefully read the "content summary" of each sub-page in the World Knowledge and analyze its relevance to the "Target Question":
\begin{enumerate}
    \item \textbf{Answer Directly (Rare)}: If the "content summary" in the World Knowledge already contains the specific factual data needed to fully answer the question, please provide the answer directly.
    \item \textbf{Explore Sub-pages (Most Common)}: If the topic of one or more sub-pages in the World Knowledge is highly relevant to the question (e.g., the question is about finding executives, and a sub-page summary is "About Us - Team Introduction"), please extract the URLs of these sub-pages and explore them to find the answer. \textbf{If you find a potential answer on these sub-pages, carefully verify its accuracy and relevance. If you are not highly confident it is the correct answer, or if you still cannot find the answer after visiting all selected sub-pages}, do NOT give up — return to the [Main Page URL] and explore it from scratch to look for additional clues or links not covered by the World Knowledge.
    \item \textbf{Explore Main Page from Scratch (Fallback)}: If all the sub-page summaries provided in the World Knowledge are completely irrelevant to the question (e.g., they are all privacy policies, disclaimers, etc.), it indicates that the current branch is invalid. In this case, you must decide to return to the [Main Page URL] to start looking for new clues from scratch.
\end{enumerate}
\end{exmp}

\end{document}